\documentclass[12pt,a4paper]{article}

\usepackage[numbers]{natbib}
\usepackage[utf8]{inputenc}
\usepackage[T1]{fontenc}
\usepackage[margin=1in]{geometry}
\usepackage{graphicx} % Images
\usepackage{amsmath}  % Equations
\usepackage{hyperref} % Clickable links
\usepackage{booktabs}
\usepackage{multirow}
\usepackage{amsfonts}
\usepackage{xcolor}
\usepackage[table]{xcolor}
\definecolor{lightred}{RGB}{255,185,185}
\definecolor{lightorange}{RGB}{255,210,150}
\definecolor{lightyellow}{RGB}{255,248,200}

\usepackage{pifont}

\title{\Large \textbf{CRISP: Progressive Illumination Curriculum for Recurrent Low-Light Video Enhancement with Semantic Priors} } 

\author{\normalsize Ruirui Lin$^*$, Xin Yan Lim$^*$, Yini Li, Zhiming Liu,  Nantheera Anantrasirichai \\  \normalsize University of Bristol \\ {\small $^*$Equal contribution} }
\date{}
\begin{document}
\maketitle

% Footnote text

% For a title note without a number/mark
%\nonumnote{}

% Here goes the abstract
\begin{abstract}
Low-light video enhancement requires the recovery of brightness, color, and fine spatial structures while preserving temporal coherence in the presence of noise and motion. Existing methods often struggle to balance these objectives, particularly in severely degraded regions. This paper presents \textbf{CRISP}, a recurrent framework trained using a progressive illumination curriculum. Its two-frame recurrent architecture propagates information across successive frames to exploit temporal context beyond the immediate input pair. During training, supervision progressively transitions from intermediate targets interpolated between the low-light input and the reference to the full reference, gradually increasing the required brightness and color correction. This curriculum stabilizes recurrent training without introducing additional inference cost and yields the largest performance improvement among the components evaluated in the ablation study. To further support structural reconstruction, the framework optionally incorporates frozen DINOv3 semantic features through adaptive layer normalization (AdaLN)-style modulation, without fine-tuning the foundation model. Experiments on DID and BVI-RLV demonstrate that CRISP achieves the highest PSNR on DID and the highest SSIM on BVI-RLV among the evaluated methods, while maintaining computational efficiency.
\end{abstract}

% Main text
% ==================================================
\section{Introduction}
\label{sec:intro}

Low-light video enhancement aims to restore visually usable videos captured under insufficient illumination. Such videos often suffer from severe sensor noise, low contrast, color distortion, and motion blur, which degrade both perceptual quality and the reliability of downstream vision tasks such as detection, tracking, and recognition~\cite{wei2018retinex,guo2020zerodce}. Restoring natural appearance remains challenging because weak illumination and amplified noise frequently introduce color casts and unstable color balance after enhancement. While existing low-light image enhancement methods can improve individual frames, directly applying them frame-by-frame to video often produces temporal inconsistencies such as flickering and ghosting due to the lack of explicit temporal modeling~\cite{Zhang2021StableLLVE,lin2024bvirlv}. Recent video-based approaches alleviate these artifacts using multi-frame temporal fusion with sliding-window propagation strategies. However, practical memory and computational constraints typically restrict the temporal window size, limiting the amount of usable temporal context under severe degradation where long-range information may be required for stable restoration~\cite{liu2025rmfat,lin2026dwta}. Moreover, large displacements between non-consecutive frames are often not fully compensated, which can lead to ghosting artifacts in the restored videos. This motivates a recurrent strategy tailored to low-light video enhancement. Although recurrent propagation has long been used in video processing, many modern learning-based restoration frameworks mainly focus on latent feature propagation, with less attention to the spatial characteristics of low-light degradation~\cite{chan2022basicvsrpp, liang2022rvrt, BRVE_2024_CVPR}. In particular, low-light videos are heavily affected by spatially uncorrelated sensor noise, which can accumulate through temporal fusion.

To address these limitations, we employ a two-frame recurrent enhancement framework that propagates the previous restored frame $O_{t-1}$ to guide enhancement of the current low-light input $I_t$. Rather than aggregating temporal information through an explicit sliding window, long-term context is accumulated implicitly through recurrence, since $O_{t-1}$ depends on earlier frames. Unrolling the model therefore makes $O_t$ indirectly dependent on the preceding sequence $I_{1:t}$, with the influence of earlier frames gradually attenuating over time. This increases the effective temporal receptive field without increasing the per-step cost beyond that of a two-frame model. Moreover, operating on adjacent frames reduces inter-frame displacement, facilitating more reliable alignment under dynamic motion and reducing ghosting artifacts. Low-light degradation can nevertheless increase appearance differences and motion ambiguity between $I_t$ and $O_{t-1}$. Existing two-frame recurrent methods typically concatenate the two inputs before restoration \cite{10943835, liu2025rmfat}, i.e., $O_t=\mathcal{W}_\theta(\text{Concat}(I_t,O_{t-1}))$. Although efficient, early concatenation can fuse misaligned structures before motion is adequately modeled. We therefore process $I_t$ and $O_{t-1}$ separately and formulate the restoration as $O_t = \mathcal{W}_\theta(I_t, O_{t-1})$, with temporal fusion and alignment performed within the decoder, i.e., $O_t=\mathcal{D}_\theta\left(\mathcal{E}^{I}_\theta(I_t),\mathcal{E}^{O}_\theta(O_{t-1})\right)$. This enables flexible motion-aware feature integration and reduces artifacts caused by premature fusion of misaligned information.

Beyond temporal modeling, another key challenge in learning-based low-light enhancement is optimization stability. Existing approaches typically treat all degradation types uniformly throughout training, despite the fact that severe sensor noise often dominates the degradation process and substantially deteriorates restoration performance~\cite{yi2024comprehensive}. In low-light conditions, increasing image brightness simultaneously amplifies noise, making the restoration objective highly ill-posed during early training stages. As a result, the network may struggle to learn stable enhancement mappings, leading to optimization instability and degraded reconstruction quality.  Some methods address this issue by explicitly designing architectures that perform denoising prior to enhancement~\cite{Shi_2024_CVPR,li2026tempretinex}, while others incorporate dedicated degradation-estimation modules to guide the restoration process~\cite{10810469, luo2025lowlight}.

Instead of increasing architectural complexity, we follow the principle of curriculum learning, where models benefit from learning simpler objectives before progressing to more challenging ones~\cite{bengio2009curriculum}. We propose a progressive illumination curriculum that gradually increases the restoration difficulty during training. Specifically, we interpolate the supervision target from the low-light input toward the full reference, thereby gradually reducing the residual degradation retained in the target. Rather than ranking training samples by difficulty, the schedule therefore follows the illumination degradation itself. That is, the model is first trained on targets close to the input, which require only a small fraction of the full brightness correction and therefore avoid demanding strong noise amplification early in training. The targets then move progressively towards the reference, with additional chroma supervision for color recovery, before training on the full ground truth with reconstruction, frequency-aware and color-aware objectives. The curriculum is applied only during training, and its removal results in the largest performance drop among all components evaluated in our ablation study.

We further address two important restoration
properties: color fidelity and structural coherence. To improve color fidelity, the proposed framework incorporates explicit chroma supervision in the CIELAB color space. In addition to standard reconstruction objectives, chromatic consistency is enforced through penalties on the $a^*$ and $b^*$ channels, encouraging more natural color reproduction and reducing color shifts commonly observed in low-light enhancement.
Furthermore, to improve structural coherence, we investigate the integration of frozen DINOv3 semantic priors~\cite{simeoni2025dinov3} through AdaLN-style modulation~\cite{peebles2023dit}. Although DINOv3 is not explicitly trained for low-light enhancement, its large-scale and diverse pretraining enables robust structural representations that can provide useful guidance when local visual evidence is degraded, offering cues beyond pixel-level information. By injecting these semantic priors into the bottleneck, decoder, and refinement stages, the proposed framework encourages more structurally coherent restoration, improved object boundary preservation, and perceptually more stable enhancement in severely degraded regions.

In summary, the main contributions are as follows:
\begin{itemize}
    \item We propose \textbf{CRISP}, \textbf{C}urriculum-guided \textbf{R}ecurrent \textbf{I}llumination Enhancement with \textbf{S}emantic \textbf{P}riors, which combines two-frame temporal propagation, internal feature alignment, and structural conditioning for efficient and temporally coherent restoration.
\item We introduce a progressive illumination curriculum that guides recurrent low-light video enhancement from input-anchored intermediate targets to full brightness, noise and color restoration, substantially improving restoration quality and stability.
\item We integrate frozen DINOv3 structural priors and CIELAB chroma supervision as complementary guidance for preserving scene structure and color fidelity, and demonstrate through extensive experiments and ablations that the resulting framework achieves competitive performance across multiple low-light video benchmarks.
\end{itemize}

% ========================================================
\section{Related Work}

\subsection{Low-Light Image Enhancement}
Early methods relied on hand-crafted techniques such as histogram equalization and Retinex-based decomposition~\cite{land1977retinex} to improve contrast and visibility. These methods often amplify noise and produce unstable color reproduction under severe degradation. Deep learning shifted the field towards learned models that jointly perform denoising, contrast recovery and color correction, including Retinex-inspired decomposition networks~\cite{zhang2019kind}, multi-branch CNNs~\cite{lv2018mbllen} and Transformer-based restorers~\cite{zamir2022restormer}. Zero-reference learning~\cite{guo2020zerodce} removes the need for paired data, while diffusion models have been combined with Retinex and semantic priors for improved realism~\cite{wu2024joresdiff}; foundation-model priors are discussed in Section~2.3. However, when applied frame by frame to video, these methods lack temporal modeling and produce flickering and inconsistent color between frames~\cite{Zhang2021StableLLVE,lin2024bvirlv}.

\subsection{Low-Light Video Enhancement}
A key challenge of video restoration is temporal alignment, commonly addressed using optical flow or latent feature alignment. EDVR~\cite{wang2019edvr} introduced deformable alignment, while recurrent frameworks including BasicVSR and BasicVSR++~\cite{chan2021basicvsr,chan2022basicvsrpp,liang2022rvrt} demonstrated the effectiveness of bidirectional propagation and feature reuse. Inaccurate alignment, however, introduces structured artifacts, and low-light video aggravates this because severe noise, blur and color distortion make correspondence estimation unreliable. Low-light-specific methods therefore couple alignment with enhancement in different ways. PCDUNet~\cite{lin2024bvirlv} and STA-SUNet~\cite{lin2024stasunet} perform explicit feature alignment within U-shaped restoration networks. DP3DF~\cite{xu2023dp3df} predicts parametric 3D filters for joint denoising and illumination enhancement, and LLVE-STCD~\cite{xu2025low} enforces spatial-temporal consistency through a decomposition-based formulation. Liu et al.~\cite{liu2023event} synthesize events from consecutive frames and use them as motion cues to guide enhancement, while DWTA-Net~\cite{lin2026dwta} adaptively weights temporal aggregation under extreme noise. Generative and sequence models have also been adopted, including conditional diffusion with wavelet interscale attention~\cite{sun2024bvicdm}, visual state-space models~\cite{huang2025bvimamba,gu2023mamba} and unsupervised Retinex-based learning under diverse lighting~\cite{li2026tempretinex}. Most of these methods fuse a fixed sliding window of frames, so the usable temporal context is bounded by memory and computation. Recent methods adopt recurrent, transformer-based, or state-space architectures~\cite{liu2025rmfat,gu2023mamba}. WaveLUT~\cite{He2026waveLUT} leverages wavelet-domain lookup priors for adaptive enhancement. The reliability-aware temporal fusion has been introduced to reduce artifacts in regions with fast motion, occlusion, or severe degradation~\cite{wang2019edvr,chan2022basicvsrpp}.

In contrast to sliding-window fusion, CRISP adopts two-frame output recurrence, so temporal context accumulates without enlarging the input window, and improves optimization through a training-only illumination curriculum rather than additional architectural modules.

\subsection{Semantic Priors in Restoration}
Recent research has shown that high-level semantic information can provide useful guidance for image enhancement. Most approaches incorporate scene-level priors into restoration pipelines~\cite{johnson2016perceptual,wu2023semanticaware}. This has expanded further through the foundation models, including SAM, CLIP, and the DINO family, whose pre-trained representations provide transferable structural and perceptual information for restoration tasks~\cite{zhang2024samir,radford2021clip,wang2023clipiqa,caron2021dino,oquab2023dinov2,simeoni2025dinov3}. Semantic priors have also been integrated with Retinex and diffusion-based low-light enhancement~\cite{wu2024joresdiff}.
These priors are typically incorporated through conditioning mechanisms such as cross-attention, spatial feature transformation, and adaptive normalization~\cite{wang2018sft,park2019spade}. The underlying motivation is that semantic information can help preserve scene structure, and perceptual coherence when pixel-level evidence is degraded.
This is particularly relevant for extreme low-light enhancement, where weak signals and heavy noise often make local restoration unreliable. Although semantic priors cannot fully solve the restoration problem, they can bias the model toward more structurally plausible and temporally stable outputs, making them attractive auxiliary guidance for low-light video enhancement.

\subsection{Curriculum Learning in Restoration.}
Curriculum learning refers to a training strategy in which task difficulty is gradually increased during optimization~\cite{bengio2009curriculum}. In image restoration, this idea was first explored in~\cite{Gao_2017_ICCV}, where corruption levels were progressively redistributed throughout training. ProSR~\cite{Wang_2018_CVPR_Workshops} and SRFBN~\cite{Li_2019_CVPR} introduced the progressive pyramid reconstruction and multi-target image super-resolution, respectively. Restore from Restored~\cite{Lee_2021_CVPR} leveraged previous restored outputs as pseudo ground truth, while Ditch the Denoiser~\cite{NEURIPS2025_c6a4fc50} adopted a staged optimization strategy that first trains on denoised samples before introducing noisy observations. In low-light enhancement, ResCue~\cite{11202696} progressively samples training patches according to residual difficulty. In contrast to these strategies, which rank or sample training examples by difficulty, our proposed curriculum modifies the supervision target itself along a progressive restoration path from the low-light input to the full reference.

%-------------------------------------------------------------------------

\section{Methodology}

\begin{figure*}
    \includegraphics[width=\textwidth]{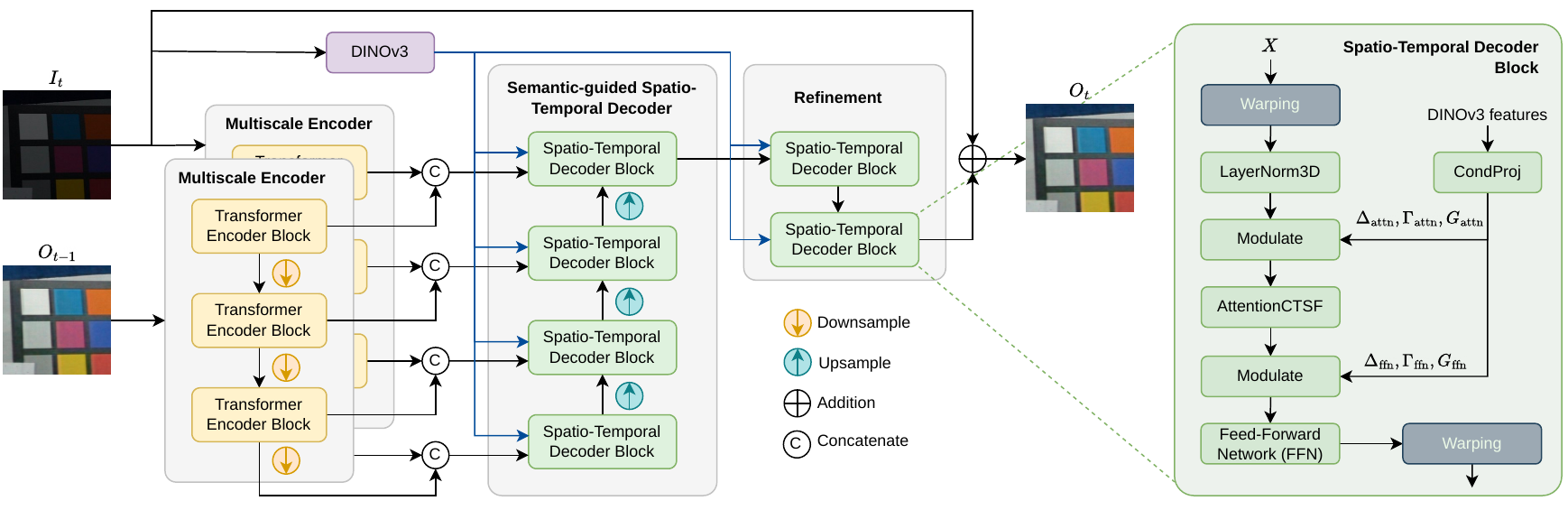}
     \smallskip
    \makebox[0.72\textwidth][c]{\small (a) Recurrent enhancement pipeline}%
    \makebox[0.27\textwidth][c]{\small (b) Decoder block}
    
    \caption{\textit{Overview of \textbf{CRISP}.
    (a) The recurrent enhancement pipeline combines framewise
    spatial encoding, multi-scale temporal decoding, and frozen
    DINOv3 semantic guidance.
    (b) Expanded view of the semantic-guided spatio-temporal
    decoder block, incorporating feature warping, CTSF attention,
    and DINOv3-guided AdaLN modulation.}} 
    \label{fig:master_pipeline} 
\end{figure*}

\autoref{fig:master_pipeline} presents an overview of the proposed CRISP framework for low-light video enhancement. The framework follows a recurrent two-frame formulation, where at each time step $t$, the model takes the current low-light frame $I_t$ together with the previous restored output $O_{t-1}$ and predicts the restored frame $O_t$. The architecture employs a multiscale encoder-decoder design with DINOv3-guided semantic conditioning injected into the decoder. Recursive propagation is applied at the output stage to enable temporally coherent enhancement across video frames. Architecture and optimization details are provided below.

%-------------------------------------------------------------------------

% ------------------------------------------------
\subsection{Multiscale Encoder}

The encoder extracts spatial features independently from the
current low-light frame $I_t$ and the previous restored output
$O_{t-1}$ before spatio-temporal processing. Since these inputs
can differ substantially in appearance and spatial alignment,
framewise encoding preserves their individual representations
before temporal integration. Direct early spatio-temporal fusion using 3D operations may therefore propagate misaligned structures and introduce ghosting artifacts, particularly under fast dynamic motion.

In implementation, the
two-frame input is reshaped from
$\mathbb{R}^{B \times 3 \times 2 \times H \times W}$ to
$\mathbb{R}^{2B \times 3 \times H \times W}$, allowing the
encoder to process both frames without mixing temporal
information.

A $3 \times 3$ convolution first projects each RGB frame
into $C$ feature channels. The resulting features pass through
three encoder levels containing 2, 3, and 3 Transformer
blocks, respectively, with 1, 2, and 4 attention heads.
The feature dimensions at these levels are
$C \times H \times W$,
$2C \times H/2 \times W/2$, and
$4C \times H/4 \times W/4$.
Between consecutive levels, a $3 \times 3$ convolution
followed by pixel unshuffle reduces each spatial dimension
by a factor of two while doubling the channel width. Compared with convolutional layers, Transformer-based feature aggregation behaves similarly to non-local denoising by modeling long-range feature interactions~\cite{10889336}.

Each Transformer block follows a Restormer-style design~\cite{zamir2022restormer},
comprising channel-wise multi-head attention and a gated
feed-forward network, both preceded by layer normalization
and connected through residual additions:
\begin{align}
    \widetilde{F}
    &= F + \mathcal{A}\bigl(\operatorname{LN}(F)\bigr), \\
    F'
    &= \widetilde{F}
    + \mathcal{G}\bigl(\operatorname{LN}(\widetilde{F})\bigr),
\end{align}
where $\mathcal{A}$ and $\mathcal{G}$ denote the attention
and feed-forward modules, respectively. The attention
module generates queries, keys, and values using a
$1 \times 1$ convolution followed by a depthwise
$3 \times 3$ convolution. Queries and keys are normalized
along the spatial dimension, and attention is computed
between channels within each head. This aggregates
image-wide information to model channel dependencies
without constructing a spatial attention matrix, yielding
linear complexity in the number of pixels for a fixed
channel width.

The feed-forward module uses a $1 \times 1$ projection
and a depthwise $3 \times 3$ convolution to produce two
feature branches with an expansion factor of 2.66.
One branch is activated by GELU and multiplied
elementwise with the other, after which a final
$1 \times 1$ projection restores the original channel
dimension. This introduces local spatial mixing and
multiplicative feature gating, suppressing irrelevant features. 

At each encoder level, the framewise features are
reshaped back into a two-frame tensor,
$F^{(l)} \in
\mathbb{R}^{B \times C_l \times 2 \times H_l \times W_l}$.
The three feature tensors provide skip connections to
the corresponding decoder levels. The coarsest tensor
also passes through a 3D convolution and spatial pixel
unshuffle to form an $8C$-channel representation at
one-eighth resolution for the spatio-temporal bottleneck.
Thus, temporal interaction begins after the three
framewise encoder levels, while their multiscale
features remain available for subsequent reconstruction.

% ------------------------------------------------
\subsection{Semantic-guided Spatio-Temporal Decoder}

The decoder combines the independently encoded features of the current low-light frame and the previous restored output through motion alignment and semantic-guided temporal fusion. Feature warping establishes temporal correspondence, while frozen DINOv3 features modulate the decoder to support structural reconstruction. We describe these two components below.

% ------------------------------------------------
\subsubsection{Temporal Feature Alignment via Warping}

The decoder operates in a 3D spatio-temporal manner, using the previous restored frame to guide enhancement of the current frame. To handle inter-frame motion, warping blocks align encoder features before decoding, followed by a second alignment stage before fusion with coarser decoder features. This progressive double-warping strategy improves temporal correspondence and reduces ghosting under dynamic motion.
As the framework uses only two successive frames, inter-frame displacement is smaller than in multi-frame approaches. Thus, lightweight 3D convolutional warping blocks can perform feature alignment, avoiding the computational cost of deformable convolutions typically used for multi-frame fusion.

Given an intermediate feature tensor $X \in \mathbb{R}^{B \times C \times T \times H \times W}$ containing both $X_{t-1}$ and $X_t$, the warping module $g(\cdot)$ predicts a two-channel flow field:
$V_{t-1 \rightarrow t} = g(X)$, where $V \in \mathbb{R}^{B \times 2 \times T \times H \times W}$.
The module $g(\cdot)$ consists of stacked 3$\times$3$\times$3 convolutions, 3D layer normalisation, and ReLU activations, followed by a final 3$\times$3$\times$3 convolution producing horizontal and vertical motion offsets. To suppress unstable alignment, the predicted displacement values are clamped to the range $[-10,10]$ before sampling.
The aligned feature is then obtained as
\begin{equation}
\tilde{X}_t(p) = X_{t-1}\big(p + V_{t-1 \rightarrow t}(p)\big),
\end{equation}
where $p=(x,y)$ denotes a spatial coordinate. In practice, differentiable grid sampling enables sub-pixel feature alignment while preserving end-to-end training.

% ------------------------------------------------
\subsubsection{Semantic-Local Feature Fusion} 

To improve structural restoration under severe low-light degradation, the proposed framework incorporates semantic conditioning using a frozen DINOv3 feature extractor. This branch provides high-level semantic priors that guide deep restoration features while preserving the low-level texture, color, and intensity for reconstruction. Specifically, DINOv3 is applied independently to each frame, and the resulting semantic maps are reassembled along the temporal dimension before being injected into the decoder. Thus, DINOv3 supplies dense, frame-wise spatial semantic context, whereas temporal coherence is mainly handled by the CTSF attention and feature-warping. A pretrained DINOv3 ViT-Small backbone is employed and kept frozen throughout training to ensure stable semantic conditioning during curriculum optimization. Following the analysis in~\cite{lin2026pixie}, intermediate semantic features $D_s$ are extracted from layers $s \in \{2,5,8,11\}$, forming a multi-scale semantic hierarchy that captures both local structural information and high-level scene semantics without requiring features from every Transformer layer. These semantic features are injected into a hierarchy of DINO-conditioned 3D Transformer blocks $l \in \{1,2,3,4\}$ through an Adaptive Layer Normalisation (AdaLN) modulation mechanism applied to both the attention and feed-forward branches. At each hierarchy level, the corresponding semantic feature $D_s$ is selected from this hierarchy and projected as:
\begin{equation}
(\Delta_{\mathrm{attn}}^{(l)}, \Gamma_{\mathrm{attn}}^{(l)}, G_{\mathrm{attn}}^{(l)},
\Delta_{\mathrm{ffn}}^{(l)}, \Gamma_{\mathrm{ffn}}^{(l)}, G_{\mathrm{ffn}}^{(l)})
=
\mathrm{CondProj}^{(l)}(D_s),
\end{equation}
where $\Delta$ and $\Gamma$ denote the shift and scale modulation parameters, respectively, while $G$ represents learnable residual gating terms. $\mathrm{CondProj}^{(l)}(\cdot)$ denotes the conditioning projection module associated with the $l$-th DINO-conditioned Transformer block. In practice, this module consists of a SiLU activation followed by a $1\times1\times1$ convolution, with an additional depthwise $3\times3\times3$ convolution applied to smooth the projected modulation features before injection into the restoration backbone.

Let $X^{(l)}$ denote the input features at hierarchy level $l$. The semantic-conditioned modulation for the attention branch is first computed as
\begin{equation}
X_{\mathrm{mod,attn}}^{(l)}
=
\mathrm{LN}(X^{(l)}) \odot
\left(1 + \Gamma_{\mathrm{attn}}^{(l)}\right)
+
\Delta_{\mathrm{attn}}^{(l)},
\end{equation}
where $\mathrm{LN}(\cdot)$ denotes layer normalisation and $\odot$ is element-wise multiplication.
The modulated features are then processed using a transposed Cross-Temporal Spatial Fusion (CTSF) attention module, followed by residual gating:
\begin{equation}
X_{\mathrm{attn}}^{(l)} = 
X^{(l)}
+
G_{\mathrm{attn}}^{(l)} \odot
\mathrm{CTSF}\left(X_{\mathrm{mod,attn}}^{(l)}\right).
\end{equation}
Unlike conventional spatio-temporal self-attention, which computes pairwise interactions across all spatial and temporal tokens, the CTSF module performs attention in a channel-temporal representation while preserving the underlying spatial structure. This formulation avoids the quadratic computational complexity associated with dense spatial attention, thereby improving scalability for high-resolution video restoration.

The resulting features $X_{\mathrm{attn}}^{(l)}$ are subsequently passed to the feed-forward branch using a second set of DINO-conditioned modulation parameters:
\begin{equation}
X_{\mathrm{mod,ffn}}^{(l)}
=
\mathrm{LN}\!\left(X_{\mathrm{attn}}^{(l)}\right)
\odot
\left(1 + \Gamma_{\mathrm{ffn}}^{(l)}\right)
+
\Delta_{\mathrm{ffn}}^{(l)}.
\end{equation}
Finally, the output of the $l$-th conditioned Transformer block is obtained as
\begin{equation}
X_{\mathrm{out}}^{(l)}
=
X_{\mathrm{attn}}^{(l)}
+
G_{\mathrm{ffn}}^{(l)} \odot
\mathrm{FFN}\!\left(X_{\mathrm{mod,ffn}}^{(l)}\right).
\end{equation}
Through this design, DINOv3 features act as high-level semantic priors during deep feature refinement, encouraging more coherent structural reconstruction and improved object boundary preservation in severely degraded regions where local image evidence is limited.

\subsubsection{Supervised Refinement and Recurrent Error Correction}
The decoder and refinement modules provide a learned mechanism for correcting errors in the recurrent input through supervised reconstruction. The previous restored output serves as temporal guidance for the current frame, rather than being reused unchanged. At each step, the decoder combines features from the previous restoration with those from the current low-light frame. A refinement module then improves the combined features before producing the restored output. These modules are trained together to match the target for each stage, allowing the network to learn to correct errors in the previous output, such as artifacts, colour shifts, and structural distortions. During training, the network uses its own previous predictions instead of the ground-truth frames, so it learns to handle imperfect inputs. This allows the network to learn how to correct errors over time, although it does not explicitly measure the reliability of previous predictions or guarantee stable performance over arbitrarily long sequences.
% ------------------------------------------------

\begin{table*}[t]
\centering
\caption{Overview of the four-stage training curriculum.
The intermediate target is
$Y_t^{(\gamma)}=I_t+\gamma (Y_t - I_t)$.
The curriculum changes training supervision only;
the inference architecture remains unchanged.}
\label{tab:curriculum}

\setlength{\tabcolsep}{5pt}
\resizebox{\textwidth}{!}{
\begin{tabular}{clcll}
\toprule
Stage & Objective & Duration & Supervision target & Active losses \\
\midrule
1 & Gradual restoration
  & 20 epochs
  & $Y_t^{(\gamma)}$, $\gamma: 0.3 \rightarrow 0.8$
  & Charbonnier + wavelet \\
2 & Brightness and color recovery
  & 30 epochs
  & $Y_t^{(\gamma)}$, $\gamma: 0.8 \rightarrow 1.0$
  & MSE + chroma \\
3 & Full-reference refinement
  & 50 epochs
  & $Y_t$
  & Charbonnier + wavelet + chroma \\
4 & Temporal consistency
  & 30 epochs
  & $Y_t$
  & Stage 3 losses + temporal \\
\bottomrule
\end{tabular}
}
\end{table*}

\subsection{Curriculum Learning Strategy}

Low-light video enhancement requires a model to recover brightness, suppress noise, and preserve structure across time. In an output-recurrent framework, these requirements are coupled. The previous prediction supplies temporal information but can also carry restoration errors into subsequent frames. 
To improve optimization stability, the network was not directly trained against the full ground-truth target $Y_t$ from the beginning. Instead, a four-stage curriculum learning strategy was employed, in which both the supervision target and optimization objective were progressively refined during training. The rationale is twofold. First, within a scene, low-light inputs captured at different illumination levels are paired with the same normal-light reference, so training directly with $\ell_1$ or $\ell_2$ losses requires visually dissimilar inputs to be mapped onto an identical target, which destabilizes optimization. Second, this instability is aggravated in recurrent architectures, where errors propagate across time. Training therefore begins with an easier intermediate target before gradually transitioning to the full restoration objective. \autoref{tab:curriculum} summarizes the supervision targets and active losses across the four training stages.

\subsubsection{Stage 1: Gradual restoration}
During the first $N_1$ epochs, the network is trained using an intermediate supervision target that lies between the low-light input frame and the corresponding ground-truth reference. Let $I_t$ denote the input frame and $Y_t$ denote the clean reference frame. The intermediate target is formulated as
\begin{equation}
Y_t^{(\gamma)} = I_t + \gamma (Y_t - I_t),
\label{eq:intermtarget}
\end{equation}
where $\gamma$ is linearly increased from $0.3$ to $0.8$ throughout Stage~1. Smaller values of $\gamma$ provide a less demanding initial reconstruction objective, as the interpolated target partially restored targets from the input, making the initial learning task easier. Since $Y_t^{(\gamma)} - I_t = \gamma\,(Y_t - I_t)$, the network initially learns only a scaled portion of the full restoration residual. Because the target retains part of the input’s noise at its original amplitude, the network is not required to amplify noise before it has learned stable enhancement features.

As $\gamma$ increases, the supervision progressively approaches the clean reference, requiring greater appearance correction and retaining less input degradation in the target. This guides the network towards the final restoration objective. The range of $\gamma$ was determined empirically to ensure stable optimization during the initial training phase.
The optimization objective used in Stage~1 is defined as
\begin{equation}
\mathcal{L}_{\mathrm{stage1}}^{(\gamma)} =
\mathcal{L}_{\mathrm{charb}}^{(\gamma)} +
\lambda_{\mathrm{dwt}}\mathcal{L}_{\mathrm{dwt}}^{(\gamma)},
\end{equation}
where $\mathcal{L}_{\mathrm{charb}}$ is the Charbonnier reconstruction loss. $\mathcal{L}_{\mathrm{dwt}}$ is the wavelet-domain loss which is defined as $||\mathrm{DWT}(\hat{O}_t)  -\mathrm{DWT}(Y_t^{(\gamma)}) ||_1$, where $\mathrm{DWT}(\cdot)$ is a single-level discrete wavelet decomposition and $\lambda_{\mathrm{dwt}}=0.02$.

% --------------------------------------------
\subsubsection{Stage 2: Brightness and color correction}
From epoch $N_1$ to $N_2$, the supervision target is progressively shifted towards the full ground-truth reference. During this stage, $\gamma$ is increased from $0.8$ to $1.0$, enabling the model to transition gradually from learning a partially brightened target to learning the true reference frame. The interpolated supervision target defined in \autoref{eq:intermtarget} is retained, with $\gamma \in [0.8,1.0]$ throughout Stage~2.

Compared with Stage~1, this stage places greater emphasis on brightness enhancement and color restoration, while maintaining stable optimization dynamics through the gradual evolution of the supervision target. The transition over $(N_2-N_1)$ epochs avoids abrupt shifts in brightness and chromatic distributions that could otherwise destabilize training. The optimization objective is defined as 
\begin{equation}
\mathcal{L}_{\mathrm{stage2}}^{(\gamma)} =
\mathcal{L}_{\mathrm{mse}}^{(\gamma)} +
\lambda_{\mathrm{chroma}}\mathcal{L}_{\mathrm{chroma}}^{(\gamma)}, 
\end{equation}
where $\mathcal{L}_{\mathrm{mse}}$ denotes the mean squared error loss, providing direct pixel-wise supervision for brightness reconstruction. To further reduce color distortion during enhancement, we introduce a chroma consistency loss $\mathcal{L}_{\mathrm{chroma}}$ that explicitly constrains the $a^*$ and $b^*$ channels in the CIELAB color space. $\mathcal{L}_{\mathrm{chroma}}$ is formulated as the $\ell_1$ distance between the predicted and reference chroma channels: 
\begin{equation}
\mathcal{L}^{(\gamma)}_{\mathrm{chroma}} = \frac{1}{N}\sum_{i=1}^{N} | \hat{C}_{ab}(i) - C_{ab}^{(\gamma)}(i) |, 
\end{equation} 
where $\hat{C}_{ab}$ and $C_{ab}$ are the predicted and reference Lab chroma channels, and $N$ is the number of pixels. The chroma loss weight was set to $\lambda_{\mathrm{chroma}}=0.005$, selected empirically from candidate values $\{0.005, 0.01, 0.02, 0.05\}$ on validation set (\autoref{ssec:setting}).

\subsubsection{Stage 3: Full-reference refinement}
From epoch $N_2$ to $N_3$, the illumination curriculum reaches the full-reference target, and the network is fine-tuned directly against the ground truth:
$Y_t^{\mathrm{target}} = Y_t$.
At this stage, the optimization objective combines three terms: robust pixel-level reconstruction, frequency-aware structural supervision, and chromatic consistency constraints:
\begin{equation}
\mathcal{L}_{\mathrm{stage3}} =
\mathcal{L}_{\mathrm{charb}} +
\lambda_{\mathrm{dwt}}\mathcal{L}_{\mathrm{dwt}} +
\lambda_{\mathrm{chroma}}\mathcal{L}_{\mathrm{chroma}},
\end{equation}
where $\lambda_{\mathrm{dwt}}=0.02$ and $\lambda_{\mathrm{chroma}}=0.005$.
This stage allows the network to refine its predictions towards the true target distribution after the recurrent dynamics and coarse enhancement behavior have stabilized. Compared with the earlier stages, optimization is now performed under full-reference supervision, enabling finer restoration of image structures, contrast, and color fidelity. The transition at epoch $N_2$ therefore acts as a practical curriculum boundary that separates the stabilization and progressive enhancement phases from the final fine-tuning stage.

\subsubsection{Stage 4: Temporal consistency} 
After epoch $N_3$, we introduce a temporal consistency constraint. Specifically, DKM~\cite{Edstedt:DKM:2023} is used to estimate dense correspondences between consecutive ground-truth frames, mapping each pixel $p$ in frame $t$ to a location $w(p)$ in frame $t$-1, with confidence $c(p)\in[0,1]$. We warp the restored frame $\hat{O}_{t-1}$ to frame $t$ using bilinear sampling and penalize the confidence-weighted absolute difference between the aligned outputs:
\begin{equation}
\begin{aligned}
&\mathcal{L}_{\mathrm{stage4}}
= \mathcal{L}_{\mathrm{stage3}} \\
&\quad + \frac{\lambda_{\mathrm{tc}}}{\sum_{p \in \Omega} c(p)}
\sum_{p \in \Omega} c(p)
\left| \hat{O}_{t}(p) - \hat{O}_{t-1}\!\left(w(p)\right) \right|,
\end{aligned}
\end{equation}
where $\Omega$ denotes the image domain and
$\lambda_{\mathrm{tc}}=\min\!\left(1,\frac{e}{10}\right)$,
where $e$ is the epoch index within Stage~4. Confidence weighting reduces the influence of unreliable correspondences. Although Stage~3 already provides good temporal consistency, Stage~4 further suppresses perceptually noticeable temporal flickering.

\subsubsection{Learning-rate schedule across stages.}
The curriculum stages are aligned with a staged learning-rate decay schedule. Specifically, $N_1$, $N_2$, $N_3$, and the total number of epochs are set to 20, 50, 100, and 130 epochs, respectively. 
We selected the stage boundaries based on validation performance during preliminary training. We identified the points where improvements became small across several validation checks, suggesting that further training at the current stage brought limited benefits. We then fixed these boundaries for all reported experiments.

During Stage~1, the learning rate is initialized at $5\times10^{-5}$ for early coarse enhancement and structural recovery. In Stage~2, the learning rate is reduced to $3\times10^{-5}$, allowing more stable optimization as the network learns increasingly non-linear mappings associated with brightness enhancement and color correction. In Stage~3, the learning rate is further decreased to $1\times10^{-5}$ for full-reference fine-tuning. Stage~4 continues training without resetting the learning rate. Operating at a smaller step size during Stages~3 and~4 helps stabilize the recurrent optimization process and prevents overshooting as the model converges towards the target distribution.

Overall, the proposed curriculum learning strategy enables stable, progressive optimization under severe low-light degradation. Rather than learning the full restoration mapping directly, the network first reconstructs an easier intermediate target, then gradually incorporates stronger brightness and chromatic supervision before converging under full ground-truth supervision.

%-------------------------------------------------------------------------
\begin{table*}[t]
    \caption{Quantitative comparison ({PSNR $\uparrow$}, {SSIM $\uparrow$} and {LPIPS $\downarrow$}) on the DID and BVI-RLV test sets. Best, second-best, and third-best results are highlighted in red, orange, and yellow, respectively. $^*$DID results may differ from previous works, as we use the provided HD videos, while some works use the 4K versions. $^\dagger$Results reported in the original publication}
    \centering
    \resizebox{\textwidth}{!}{
    \begin{tabular}{l|ccc|ccc|ccc}
        \toprule
        \multirow{2}{*}{\textbf{Method}} 
        & \multicolumn{3}{c|}{\textbf{DID}$^*$} 
        & \multicolumn{3}{c|}{\textbf{BVI-RLV}} 
        & \multicolumn{3}{c}{\textbf{Average}} \\
        \cline{2-10}
        & \textbf{PSNR $\uparrow$} & \textbf{SSIM $\uparrow$} & \textbf{LPIPS $\downarrow$}
        & \textbf{PSNR $\uparrow$} & \textbf{SSIM $\uparrow$} & \textbf{LPIPS $\downarrow$}
        & \textbf{PSNR $\uparrow$} & \textbf{SSIM $\uparrow$} & \textbf{LPIPS $\downarrow$} \\
        \midrule

        SMOID-net~\cite{Jiang:learn:2019} 
        & 21.71 & 0.880 & 0.194 
        & 21.82 & 0.780 & 0.280 
        & 21.76 & 0.830 & 0.237 \\

        SDSD-net~\cite{wang2021sdsd} 
        & 21.88 & 0.834 & 0.216 
        & 20.60 & 0.658 & 0.148 
        & 21.24 & 0.746 & 0.182 \\

        PCDUNet~\cite{lin2024bvirlv} 
        & 18.91 & 0.725 & 0.305 
        & 24.51 & 0.845 & 0.192 
        & 21.71 & 0.785 & 0.249 \\

        STA-SUNet~\cite{lin2024stasunet} 
        & 19.61 & 0.714 & 0.385 
        & 26.14 & 0.873 & 0.199 
        & 22.88 & 0.794 & 0.292 \\

        TempRetinex~\cite{li2026tempretinex} 
        & 19.27 & 0.849 & 0.314 
        & 30.27 & 0.877 & 0.327 
        & 24.77 & 0.863 & 0.321 \\

        BVI-CDM~\cite{sun2024bvicdm} 
        & 19.44 & 0.683 & 0.199 
        & 29.93 & 0.897 & \cellcolor{yellow!35}0.097 
        & 24.69 & 0.790 & 0.148 \\

        BVI-Mamba~\cite{huang2025bvimamba} 
        & 20.47 & 0.816 & 0.157 
        & \cellcolor{yellow!35}31.22 & \cellcolor{orange!35}0.912 & \cellcolor{red!35}0.071 
        & 25.85 & 0.864 & \cellcolor{orange!35}0.114 \\

        DWTA-Net~\cite{lin2026dwta} 
        & 24.27 & 0.857 & \cellcolor{red!35}0.115 
        & 28.42 & 0.891 & \cellcolor{orange!35}0.086 
        & 26.35 & 0.874 & \cellcolor{red!35}0.101 \\

        ReMATF~\cite{liu2026rematf} & 24.26 & 0.852 &  0.231 & 30.46 & 0.869 & 0.175 & 27.36 & 0.861 & 0.203 \\
        
        WaveLUT$^\dagger$~\cite{He2026waveLUT} 
        & \cellcolor{yellow!35}26.49 & \cellcolor{red!35}0.900 & - 
        & 29.21 & 0.890 & - 
        & \cellcolor{yellow!35}27.85 & \cellcolor{yellow!35}0.895 & - \\

        LLVE-STCD~\cite{xu2025low} 
        & \cellcolor{orange!35}26.93 & \cellcolor{orange!35}0.898 & \cellcolor{yellow!35}0.141 
        & \cellcolor{red!35}33.43 & \cellcolor{yellow!35}0.919 & 0.104 
        & \cellcolor{orange!35}30.18 & \cellcolor{orange!35}0.909 & 0.123 \\

        \midrule
        Ours 
        & \cellcolor{red!35}27.76 & \cellcolor{yellow!35}0.896 & \cellcolor{orange!35}0.140 
        & \cellcolor{orange!35}32.78 & \cellcolor{red!35}0.923 & 0.098 
        & \cellcolor{red!35}30.27 & \cellcolor{red!35}0.910 & \cellcolor{yellow!35}0.119 \\
        \bottomrule
    \end{tabular}
    }
    
    \label{tab:combined_results}
\end{table*}

% --------------------------------------------------
\section{Experimental results and discussion}

\subsection{Implementation}
\label{ssec:setting}
We use the official training and test splits of DID~\cite{Fu_2023_ICCV}, BVI-RLV~\cite{lin2024bvirlv} and OLiVES~\cite{olives2026}, training a separate model for each dataset on its own training split and evaluating it only on the corresponding test split. No cross-dataset training, joint training or fine-tuning is performed, so all reported results are obtained in-domain. The initial input $O_{t-1}$ is set to $I_0$. During training, fixed-length clips of $T$=30 frames were extracted. This provides a good trade-off between performance and training time. Longer clips yielded only marginal improvements while requiring substantially more memory. The 30-frame clips are also considerably longer than the multi-frame inputs used by previous methods. Each clip is processed sequentially, starting with its first low-light frame. At each step, the model uses the current frame and its previous prediction. The prediction is detached from the computation graph before being used as the next recurrent state, and the model is updated at every frame, so gradients do not propagate across steps. Ground-truth frames are used only for supervision, not as recurrent inputs. During evaluation, the state is reset for each video and updated using the model's predictions.

The videos are randomly cropped into $512\times512$ patches. Paired low-light and reference clips were augmented jointly using random horizontal and vertical flips. The restoration backbone was optimized with Adam using a batch size of one clip per GPU. The learning rate followed the four curriculum stages, with values of $5\times10^{-5}$, $3\times10^{-5}$, $1\times10^{-5}$ and $1\times10^{-5}$ for Stages~1, 2, 3 and 4, respectively. All hyperparameters, including $\lambda_{\mathrm{chroma}}$, the range of $\gamma$ and the curriculum stage boundaries, were selected on a validation set  held out from the BVI-RLV and DID training split. The test sets were used only for final evaluation.
The DINOv3 feature extractor was kept frozen and used only for semantic conditioning. 
Our experiments were run on an H100 GPU. Note that the model can also be trained on an RTX 3090, but training takes approximately five times longer.

% -----------------------------------------------
\subsection{Performance Comparison}
We evaluate the proposed method on 1920$\times$1080 videos from DID~\cite{Fu_2023_ICCV}, BVI-RLV~\cite{lin2024bvirlv} and OLiVES~\cite{olives2026}.  Comparisons are made against existing video enhancement methods, including SMOID-net~\cite{Jiang:learn:2019}, SDSD-net~\cite{wang2021sdsd}, PCDUNet~\cite{lin2024bvirlv}, STA-SUNet~\cite{lin2024stasunet}, TempRetinex~\cite{li2026tempretinex}, DWTA-Net~\cite{lin2026dwta}, BVI-CDM~\cite{sun2024bvicdm}, BVI-Mamba~\cite{huang2025bvimamba}, ReMATF~\cite{liu2026rematf}, WaveLUT~\cite{He2026waveLUT}, and LLVE-STCD~\cite{xu2025low}. Full-reference metrics (PSNR, SSIM \cite{1284395}, and LPIPS \cite{8578166}) are computed against the corresponding ground truth. We recomputed all baselines under a common evaluation protocol, except for WaveLUT, whose results were taken from the original paper as we were unable to run its code, so its LPIPS value is unavailable.

\subsubsection{Quantitative Comparison.}
\begin{figure}
    \centering
    \includegraphics[width=0.5\columnwidth]{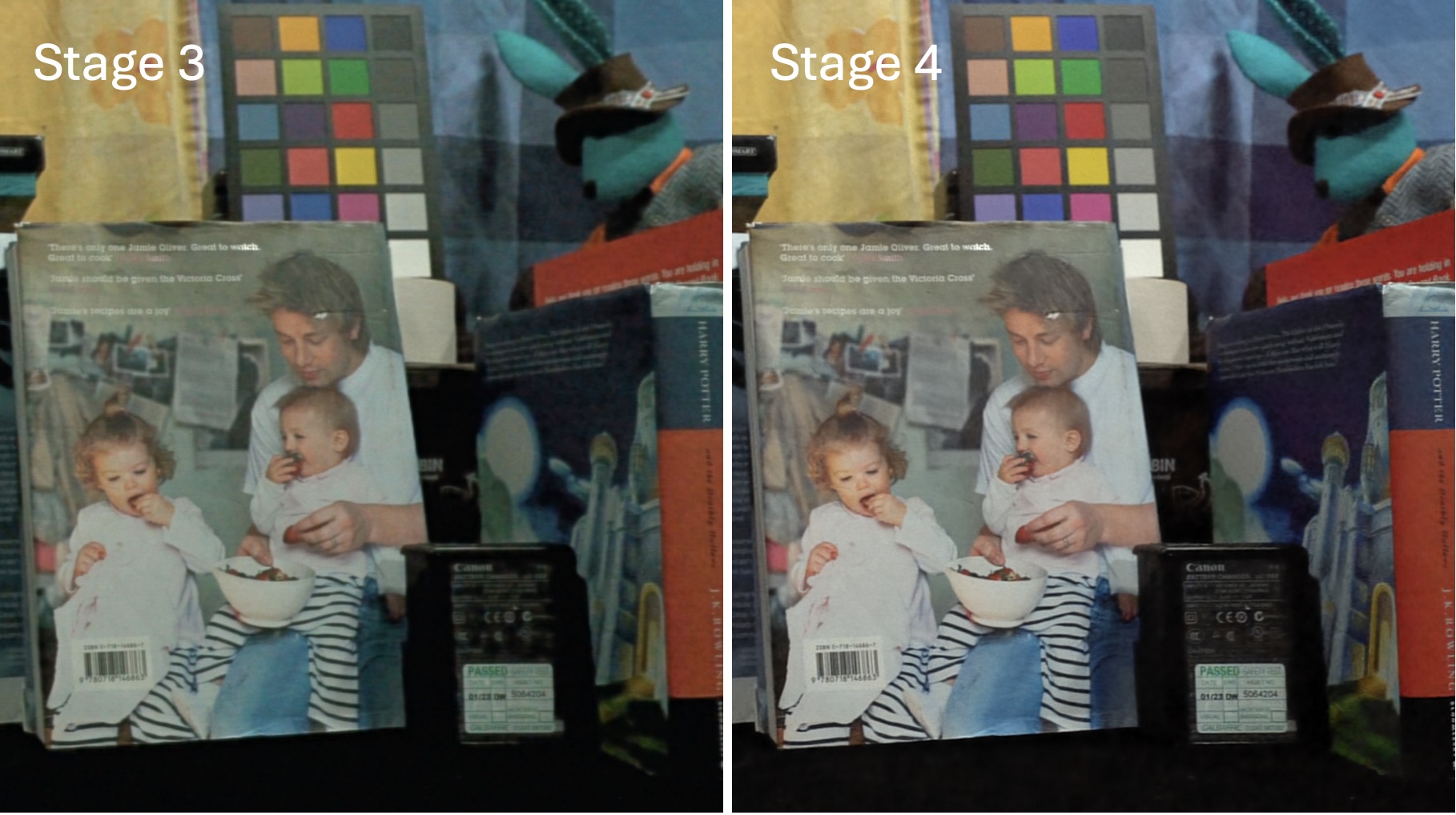}
    \caption{\textit{Spatial quality comparison between the results from Stage~3 (left) and Stage~4 (right) on the BVI-RLV dataset.} }
    \label{fig:stage3and4}
\end{figure}

As shown in \autoref{tab:combined_results}, the proposed method achieves the highest PSNR on DID and the highest SSIM on BVI-RLV, while remaining competitive across the other reconstruction metrics, demonstrating superior reconstruction fidelity and structural preservation. However, its LPIPS score remains higher than those of BVI-Mamba and DWTA-Net, indicating that while the proposed framework restores more accurate image content overall, state-space modeling in BVI-Mamba and the texture-adaptive optimization strategy of DWTA-Net may better preserve perceptual smoothness and natural appearance under the LPIPS criterion.
%We also investigated the Motion-Adaptive Temporal Fusion (MATF) module from ReMATF~\cite{liu2026rematf}, which aggregates long-term information as $O_t = M \hat{O}_t + (1-M)O_{t-1}$, where $M$ is a motion-aware weighting factor and $\hat{O}_t$ denotes the current network output. In this formulation, smaller $M$ values in static regions favor temporal aggregation from $O_{t-1}$, while larger values in fast-motion regions rely more on the current prediction. However, MATF did not improve our framework, suggesting that the decoder already learns suitable region-wise temporal fusion without explicit motion-adaptive weighting.
% -----------------------------------------------------------
For temporal consistency, we report temporal LPIPS (tLPIPS)~\cite{33865693392457} and motion-compensated warping error ($E_{\mathrm{warp}}$)~\cite{Lai_2018_ECCV}. As shown in \autoref{tab:tlpips}, before introducing temporal supervision (Stage~3), our method achieves a tLPIPS of 0.1039 and an $E_{\mathrm{warp}}$ of 0.0342. Adding the temporal consistency loss in Stage~4 substantially reduces these values to 0.0035 and 0.0207, respectively, outperforming the compared methods on both metrics. We also compare the spatial quality of the results in \autoref{fig:stage3and4}. The Stage~4 results remain sharp, indicating that the lower tLPIPS is not simply due to excessive smoothing. Note that the video was captured with camera transitions, resulting in global scene motion. These results demonstrate that Stage~4 effectively suppresses temporal inconsistencies introduced during recurrent enhancement while preserving spatial detail.
%\autoref{fig:ewarp_mix5} shows the temporal evolution of $E_{\mathrm{warp}}$ on a representative \textit{Mix5} sequence, where the proposed method tracks the ground-truth trajectory closely throughout. %Overall, the results suggest that the proposed method achieves competitive temporal consistency while maintaining strong reconstruction quality.

\begin{table*}
\centering
\caption{Temporal consistency comparison on BVI-RLV.} 
%\resizebox{\columnwidth}{!}{
\begin{tabular}{l|ccccc}
\toprule
Method & BVI-Mamba & ReMATF & LLVE-STCD & Ours (Stage3) & Ours (Stage4) \\
\midrule
tLPIPS$\downarrow$ & 0.1604 & 0.1478 & \underline{0.0336} & 0.1039 & \textbf{0.0035}  \\
$E_{warp}\downarrow$ & \underline{0.0319} & 0.0345 & 0.0349 & 0.0342 & \textbf{0.0207}\\
\bottomrule
\end{tabular}
%}
\label{tab:tlpips}
\end{table*}

\begin{figure}
    \centering
    \includegraphics[width=\columnwidth]{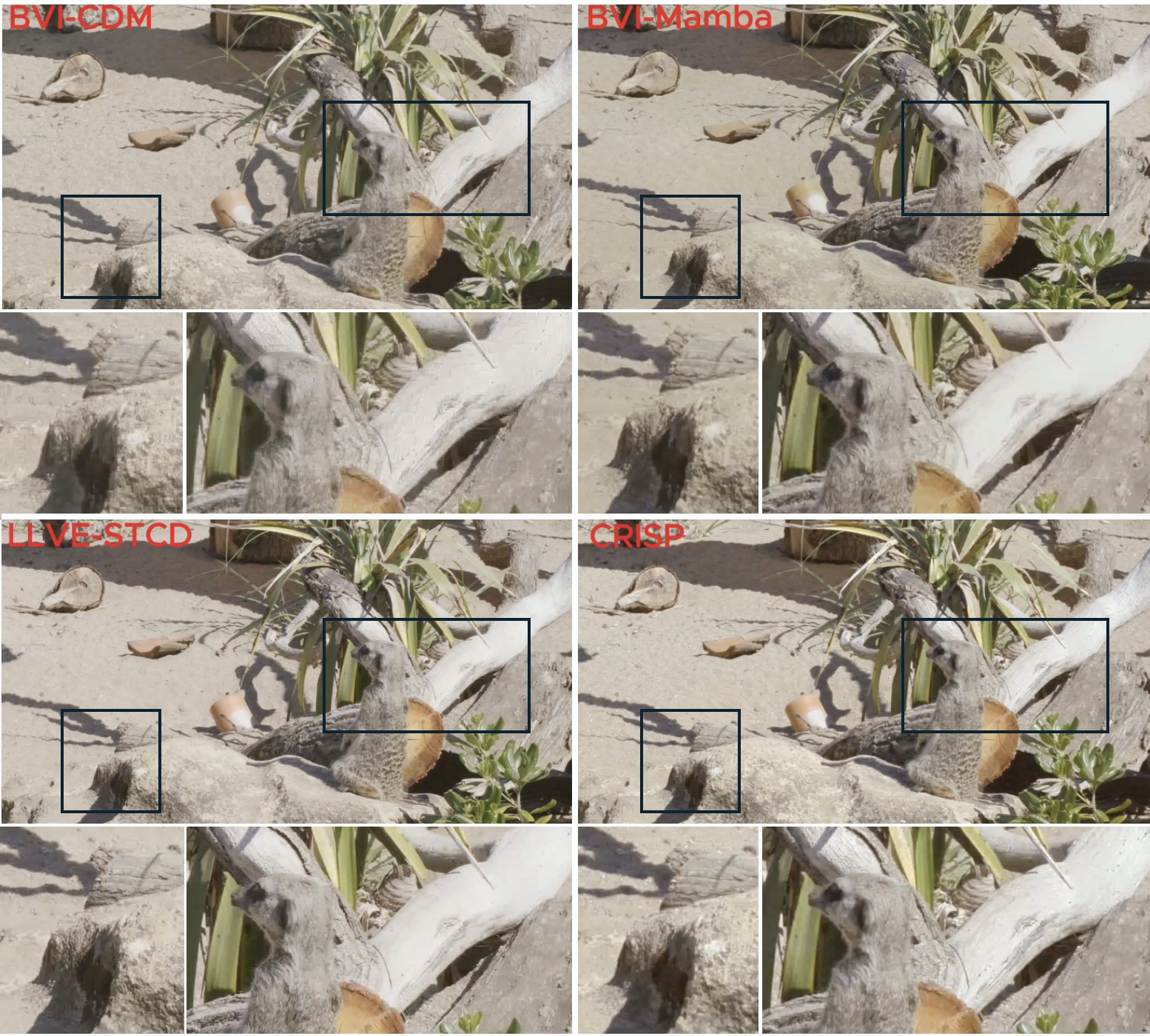}
    \caption{\textit{Qualitative comparison of the top four methods on the OLiVES dataset, as reported in \autoref{tab:OLiVES}. Clockwise from the top left: BVI-CDM, BVI-Mamba, LLVE-STCD, and ours.}}
    \label{fig:OLiVEresults}
\end{figure}

We further evaluate enhancement performance on the OLiVES dataset, which contains outdoor animal scenes with heavy noise and substantial illumination variation. \autoref{tab:OLiVES} reports baseline results provided by the OLiVES benchmark authors, while our method is evaluated independently. Although frame-level reconstruction is more challenging under these conditions, our method maintains competitive SSIM and LPIPS while achieving the lowest $E_{\rm warp}$ (0.0141), demonstrating strong motion-compensated temporal stability.  However, robustness to large illumination variations remains a limitation. \autoref{fig:OLiVEresults} shows that our CRISP method produces sharper results with finer details.

\begin{table*}
    \centering
    \caption{Performance comparison on the OLiVES dataset. Baseline results were provided by the benchmark authors, while CRISP was evaluated independently. All evaluation settings, including the data split, resolution, and metric implementation, were matched.}
   %\resizebox{0.95\columnwidth}{!}{
    \begin{tabular}{l|ccc|cc}
    \toprule
    
    {Method} & {PSNR}$\uparrow$ & {SSIM}$\uparrow$ & {LPIPS}$\downarrow$ & {tLPIPS}$\downarrow$ & {E$_{warp}$}$\downarrow$ \\
    \midrule
    SDSD-Net~\cite{wang2021sdsd} & 22.59 & 0.716 & 0.322 & 0.0600 & 0.0758 \\
    StableLLVE~\cite{Zhang2021StableLLVE} & 22.60 & 0.850 & 0.300 & 0.0280 & 0.0448 \\
    DP3DF~\cite{xu2023dp3df} & 23.22 & 0.781 & 0.325 & 0.0622 & 0.0657 \\
    BVI-CDM~\cite{sun2024bvicdm} & \cellcolor{orange!35}26.40 & \cellcolor{yellow!35}0.854 & 0.326 & \cellcolor{red!35}{0.0167} & \cellcolor{orange!35}0.0390 \\
    BVI-Mamba~\cite{huang2025bvimamba} & 25.66 & \cellcolor{red!35}{0.884} & \cellcolor{red!35}{0.228} & \cellcolor{orange!35}{0.0192} & \cellcolor{yellow!35}0.0397 \\
    LLVE-STCD~\cite{xu2025low} & \cellcolor{red!35}{27.00} & 0.827 & \cellcolor{orange!35}0.255 & 0.0575 & 0.0534 \\ \hline
    Ours & \cellcolor{yellow!35}26.38 & \cellcolor{orange!35}0.857 & \cellcolor{yellow!35}0.262 & \cellcolor{yellow!35}0.0260 & \cellcolor{red!35}0.0141\\
    
    \bottomrule
    \end{tabular}
  % }
    \label{tab:OLiVES} 
\end{table*}

% -----------------------------------------------

\subsubsection{Qualitative Comparison.}
\begin{figure*}[t]
    \centering
    \begin{minipage}[t]{0.166\textwidth}
        \centering
        \includegraphics[width=\linewidth]{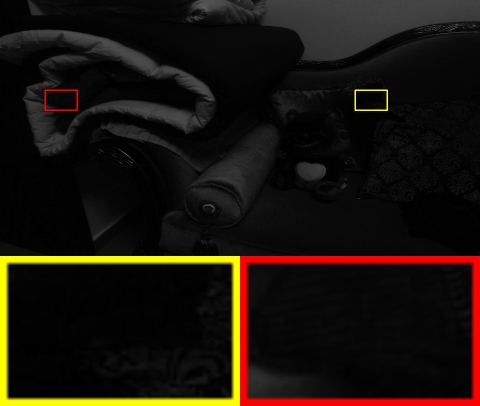}\\[-1mm]
        {\scriptsize Input}
    \end{minipage}%
    \begin{minipage}[t]{0.166\textwidth}
        \centering
        \includegraphics[width=\linewidth]{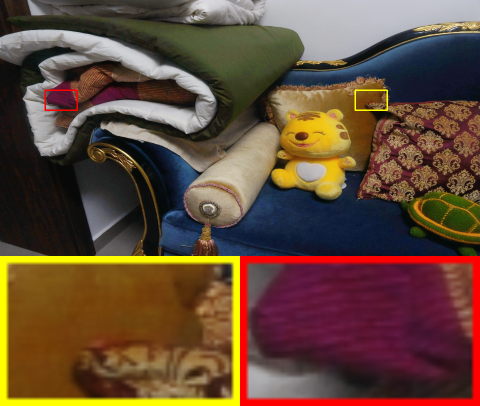}\\[-1mm]
        {\scriptsize RetinexFormer}
    \end{minipage}%
    \begin{minipage}[t]{0.166\textwidth}
        \centering
        \includegraphics[width=\linewidth]{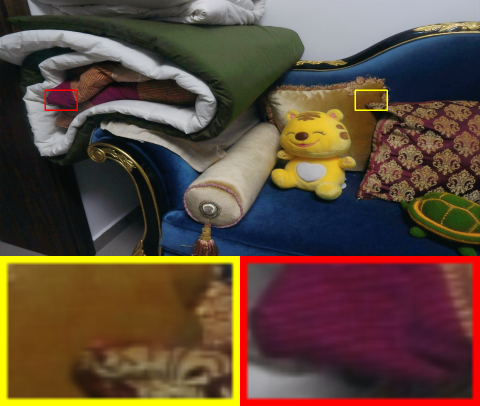}\\[-1mm]
        {\scriptsize SDSDNet}
    \end{minipage}%
    \begin{minipage}[t]{0.166\textwidth}
        \centering
        \includegraphics[width=\linewidth]{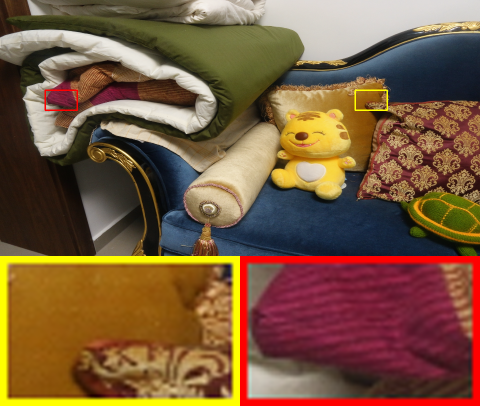}\\[-1mm]
        {\scriptsize LLVE-STCD}
    \end{minipage}%
    \begin{minipage}[t]{0.166\textwidth}
        \centering
        \includegraphics[width=\linewidth]{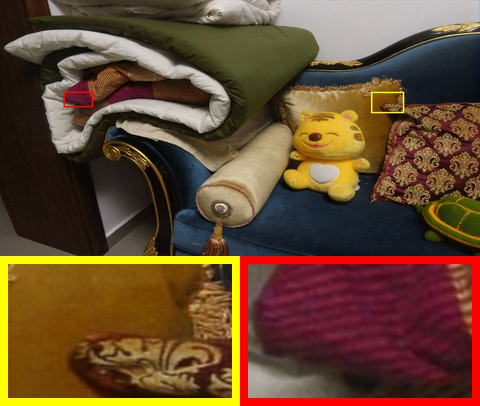}\\[-1mm]
        {\scriptsize Ours}
    \end{minipage}%
    \begin{minipage}[t]{0.166\textwidth}
        \centering
        \includegraphics[width=\linewidth]{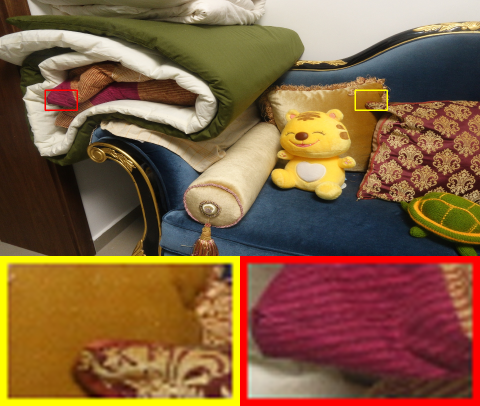}\\[-1mm]
        {\scriptsize GT}
    \end{minipage}
    \vspace{1mm}

    \centering
    \begin{minipage}[t]{0.166\textwidth}
        \centering
        \includegraphics[width=\linewidth]{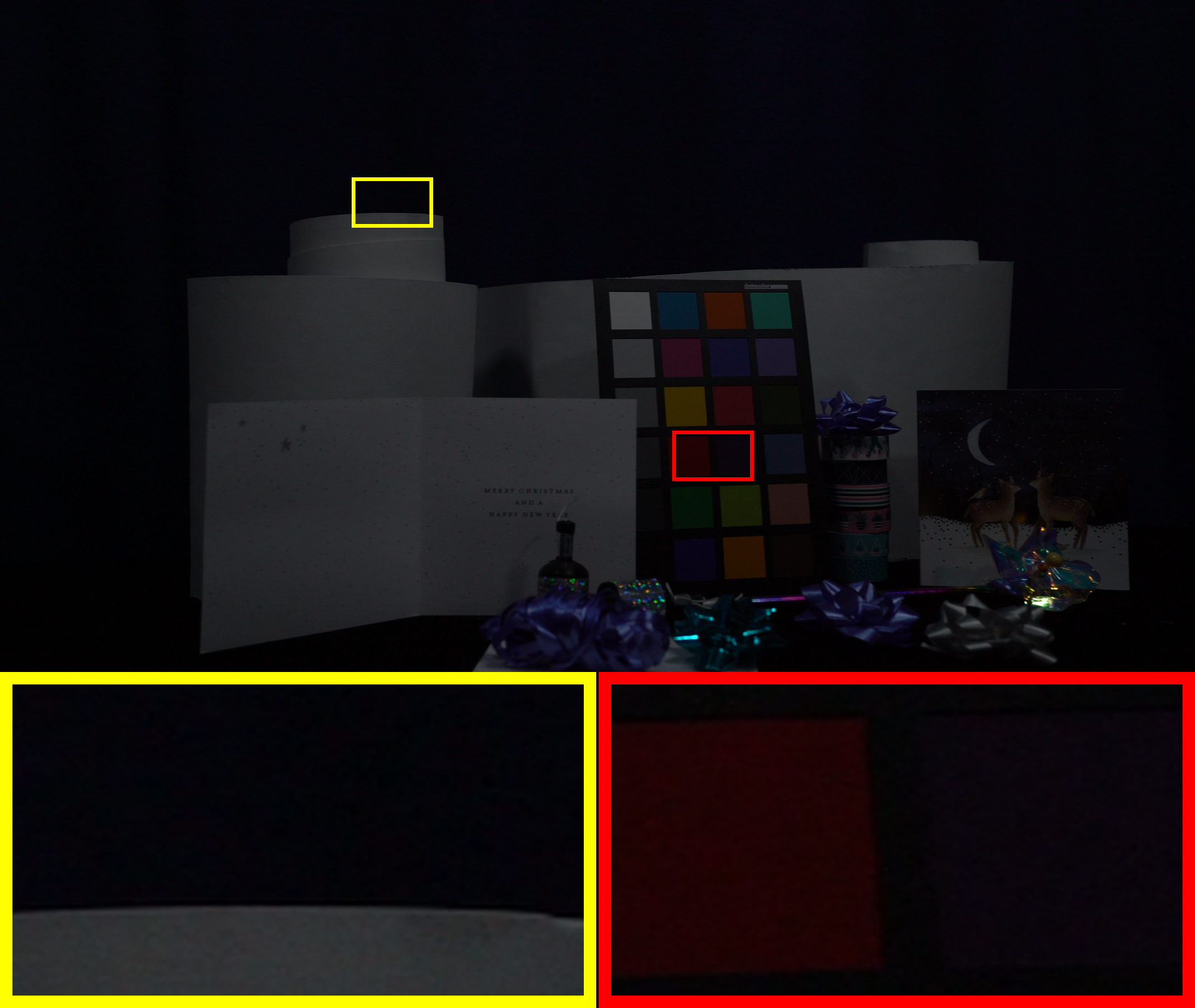}\\[-1mm]
        {\scriptsize Input}
    \end{minipage}%
    \begin{minipage}[t]{0.166\textwidth}
        \centering
        \includegraphics[width=\linewidth]{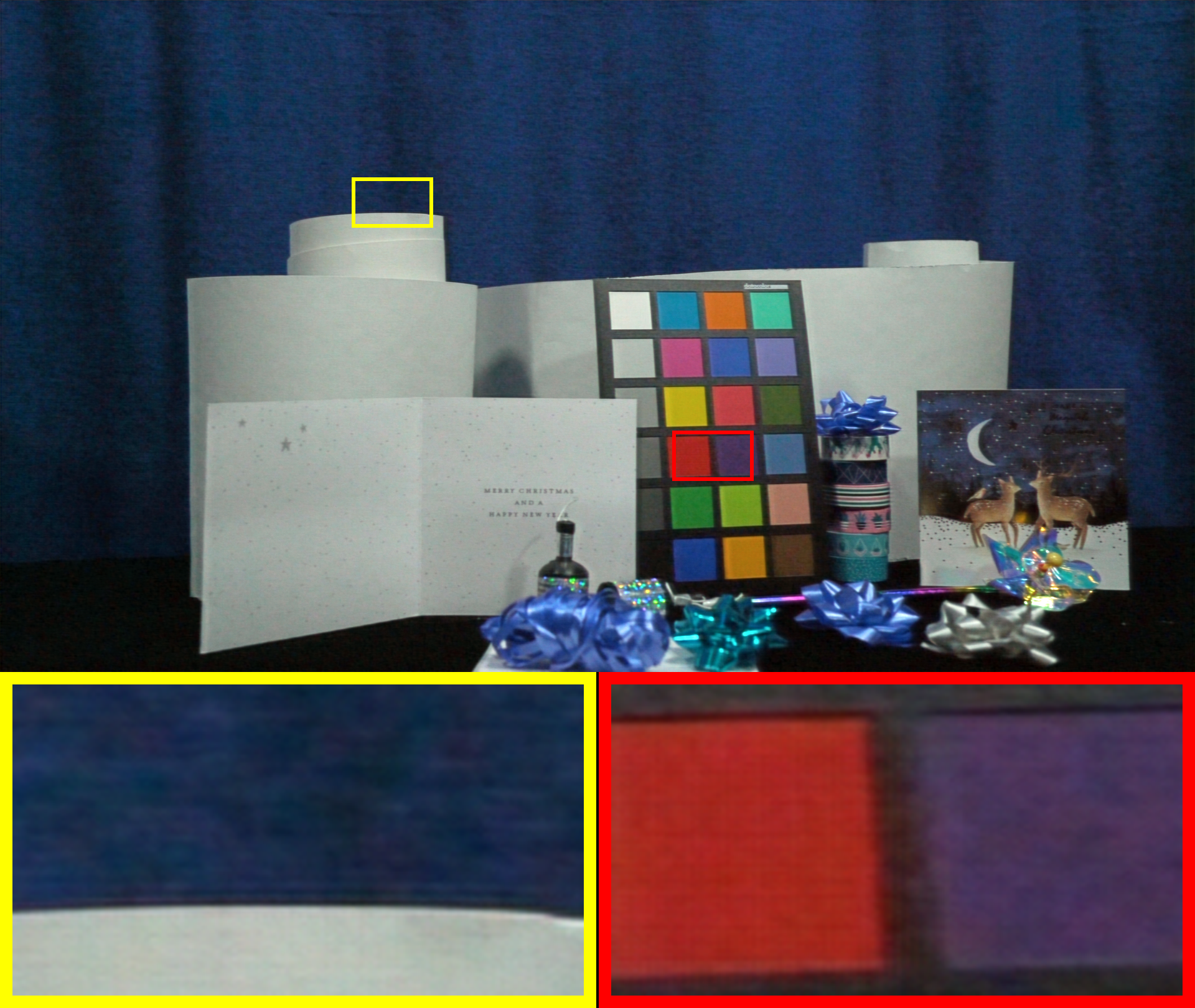}\\[-1mm]
        {\scriptsize ReMATF}
    \end{minipage}%
    \begin{minipage}[t]{0.166\textwidth}
        \centering
        \includegraphics[width=\linewidth]{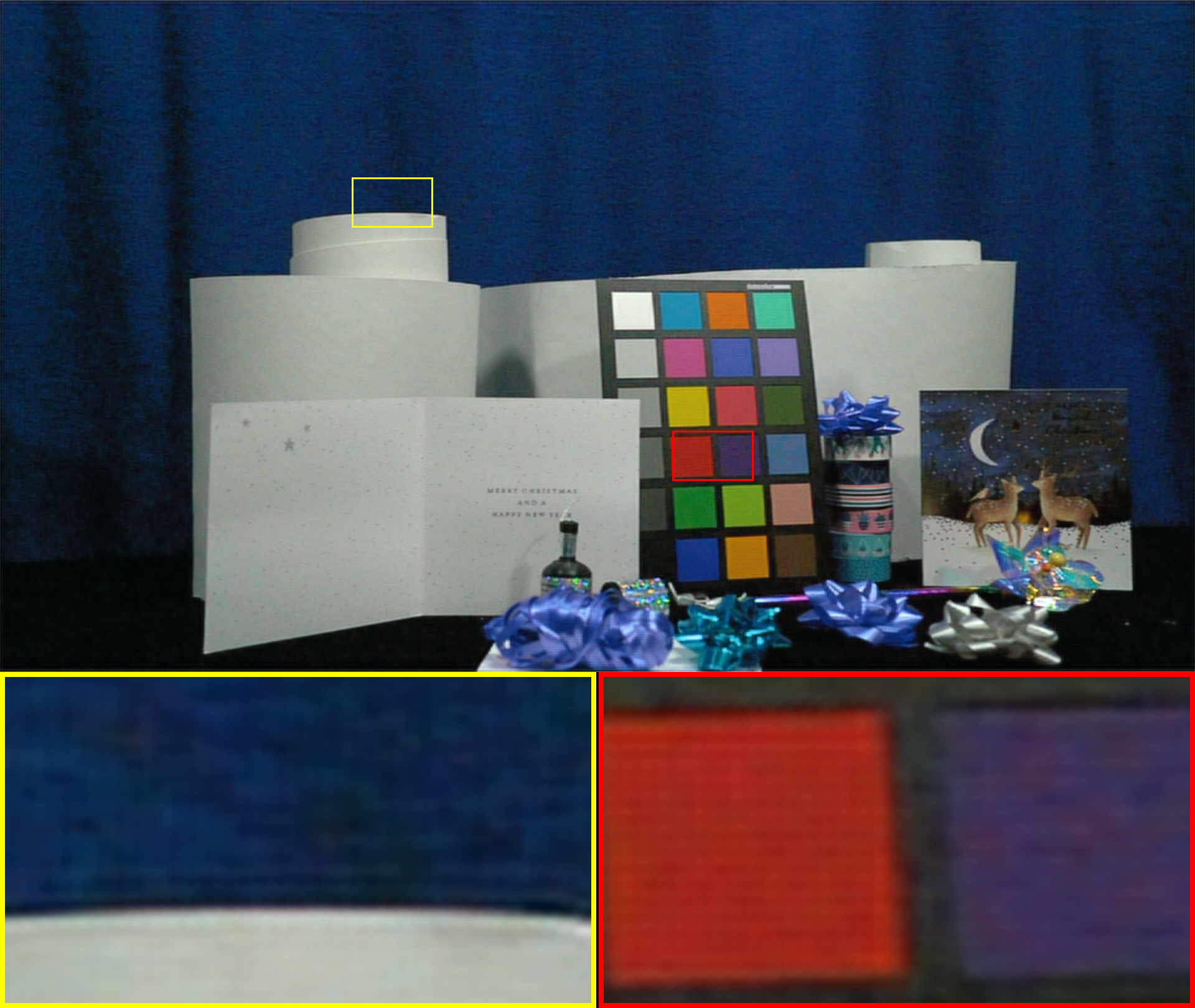}\\[-1mm]
        {\scriptsize BVI-Mamba}
    \end{minipage}%
    \begin{minipage}[t]{0.166\textwidth}
        \centering
        \includegraphics[width=\linewidth]{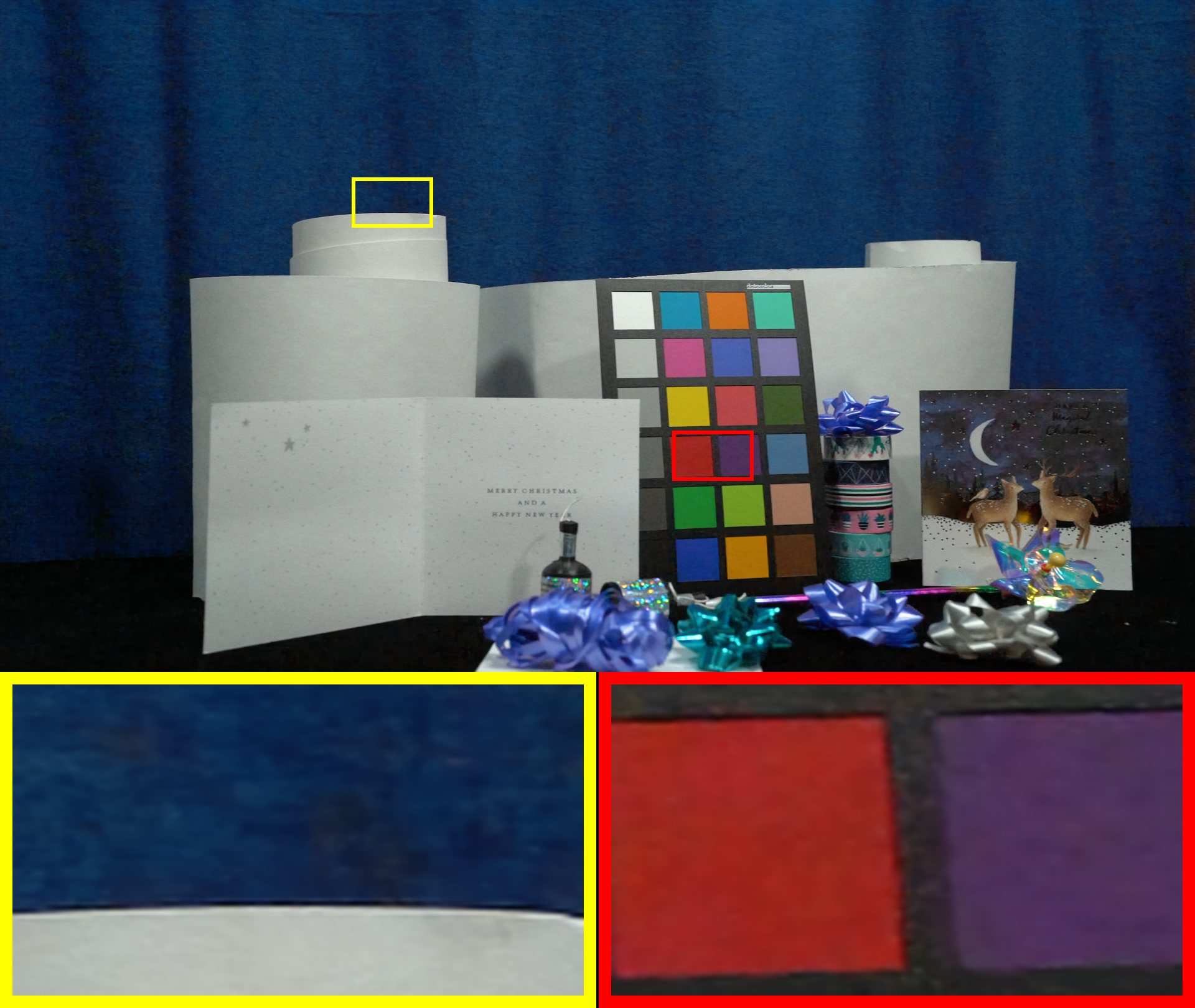}\\[-1mm]
        {\scriptsize LLVE-STCD}
    \end{minipage}%
    \begin{minipage}[t]{0.166\textwidth}
        \centering
        \includegraphics[width=\linewidth]{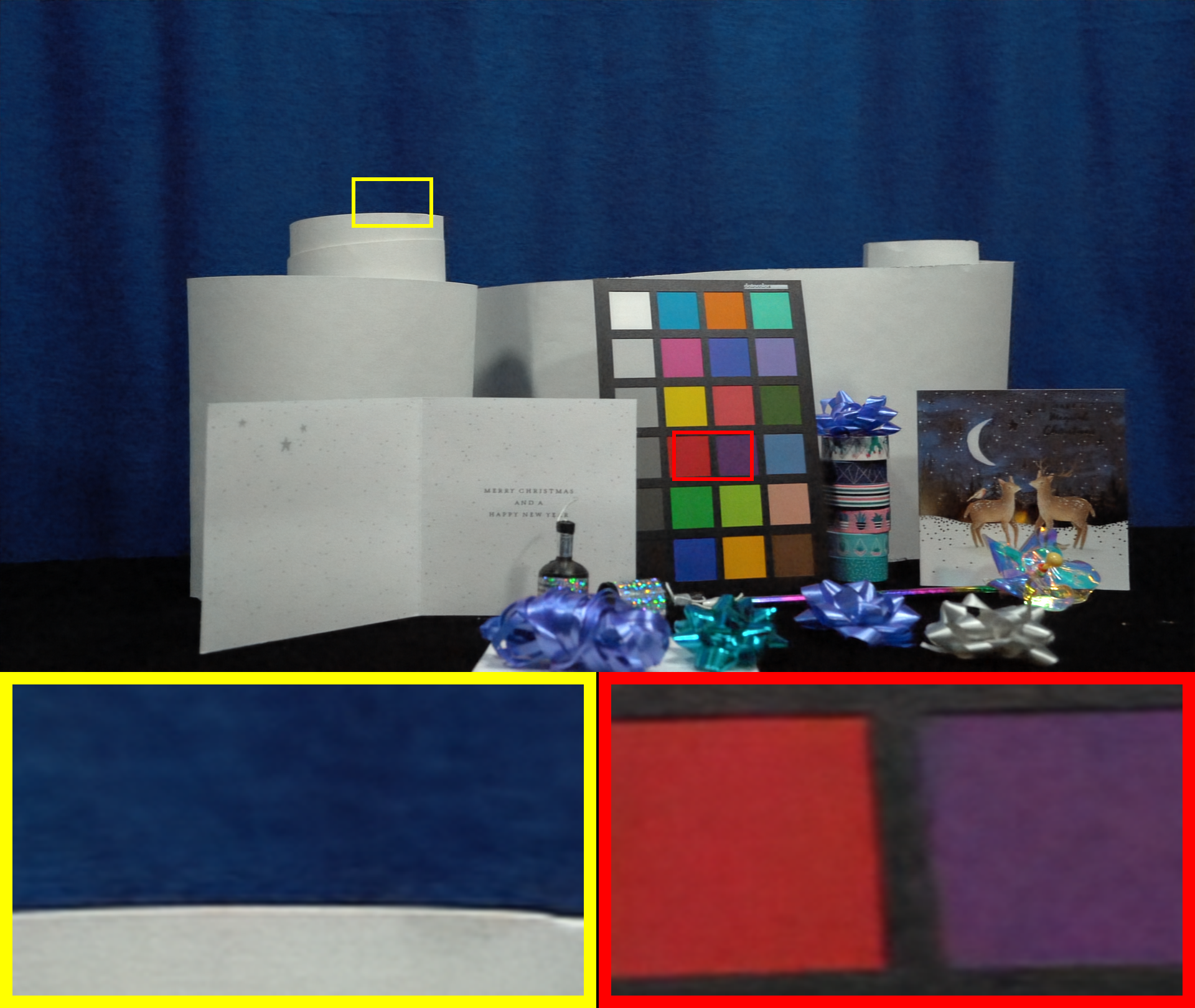}\\[-1mm]
        {\scriptsize Ours}
    \end{minipage}%
    \begin{minipage}[t]{0.166\textwidth}
        \centering
        \includegraphics[width=\linewidth]{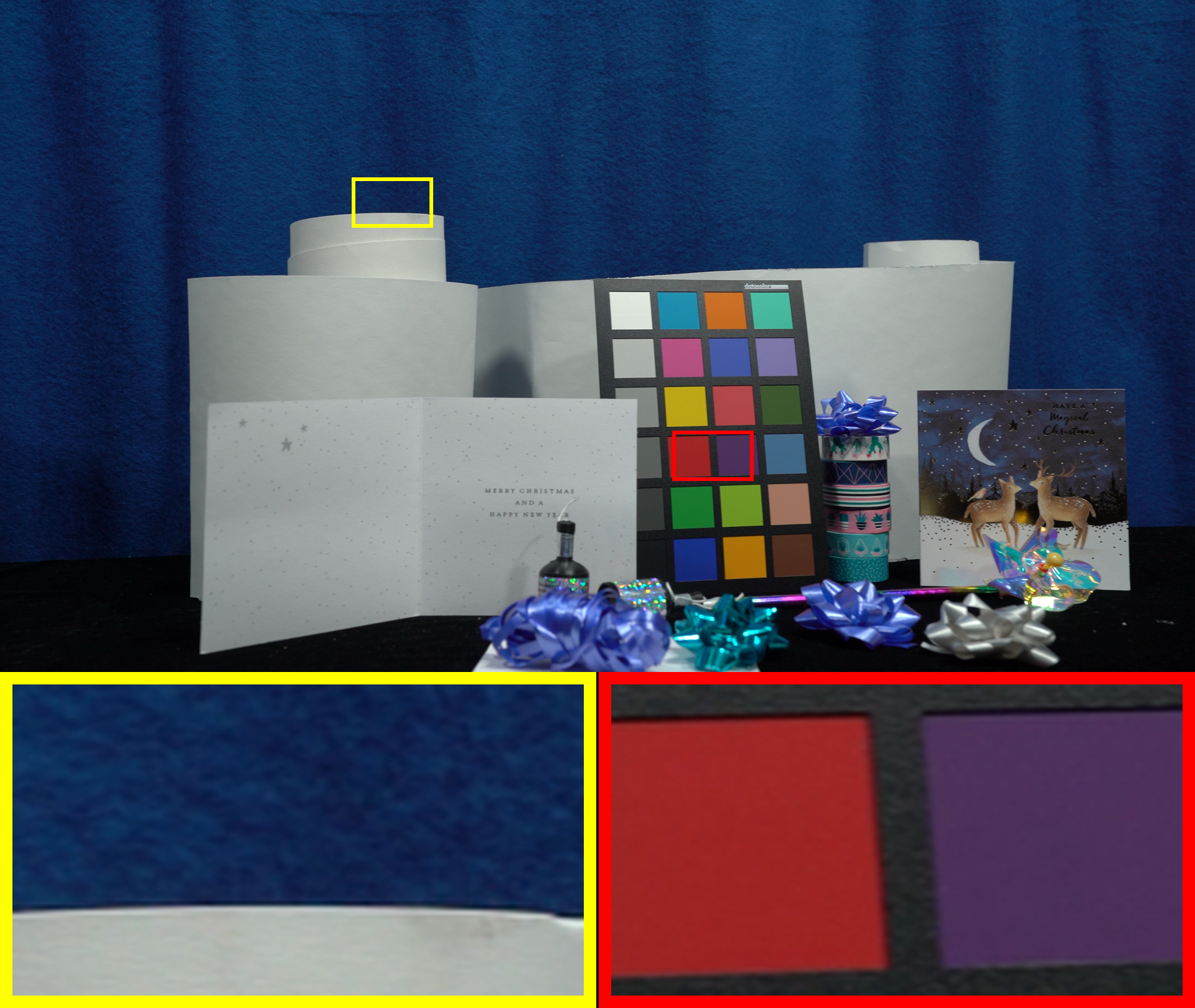}\\[-1mm]
        {\scriptsize GT}
    \end{minipage}
    %\vspace{4mm}
    \caption{\textit{Qualitative comparison on the DID (top row) and BVI-RLV (bottom row) datasets}.} 
    \label{fig:BVIsubjectiveresults}
\end{figure*}

Representative qualitative results on the DID and BVI-RLV datasets are shown in \autoref{fig:BVIsubjectiveresults}. On the DID dataset, RetinexFormer~\cite{Cai_2023_ICCV} and SDSD-net improve global brightness but soften local textures, while LLVE-STCD provides stronger enhancement with improved color and contrast. In contrast, the proposed method restores visibility while better preserving local color and fine texture, closely matching the ground truth. On the BVI-RLV dataset, ReMATF and BVI-Mamba enhance global illumination at the cost of local fidelity, while LLVE-STCD again provides strong enhancement and color recovery. The proposed method produces a more conservative reconstruction with improved denoising in homogeneous regions. This behavior is consistent across the highlighted regions: in the yellow box, our method better preserves the smooth background and boundary structures with fewer artifacts, while in the red box, it achieves color recovery comparable to LLVE-STCD while retaining clearer structural details. Although LLVE-STCD produces a global illumination level closer to the ground truth, our reconstructions are consistently sharper. 

\begin{figure}
    \centering
    \includegraphics[width=0.75\columnwidth]{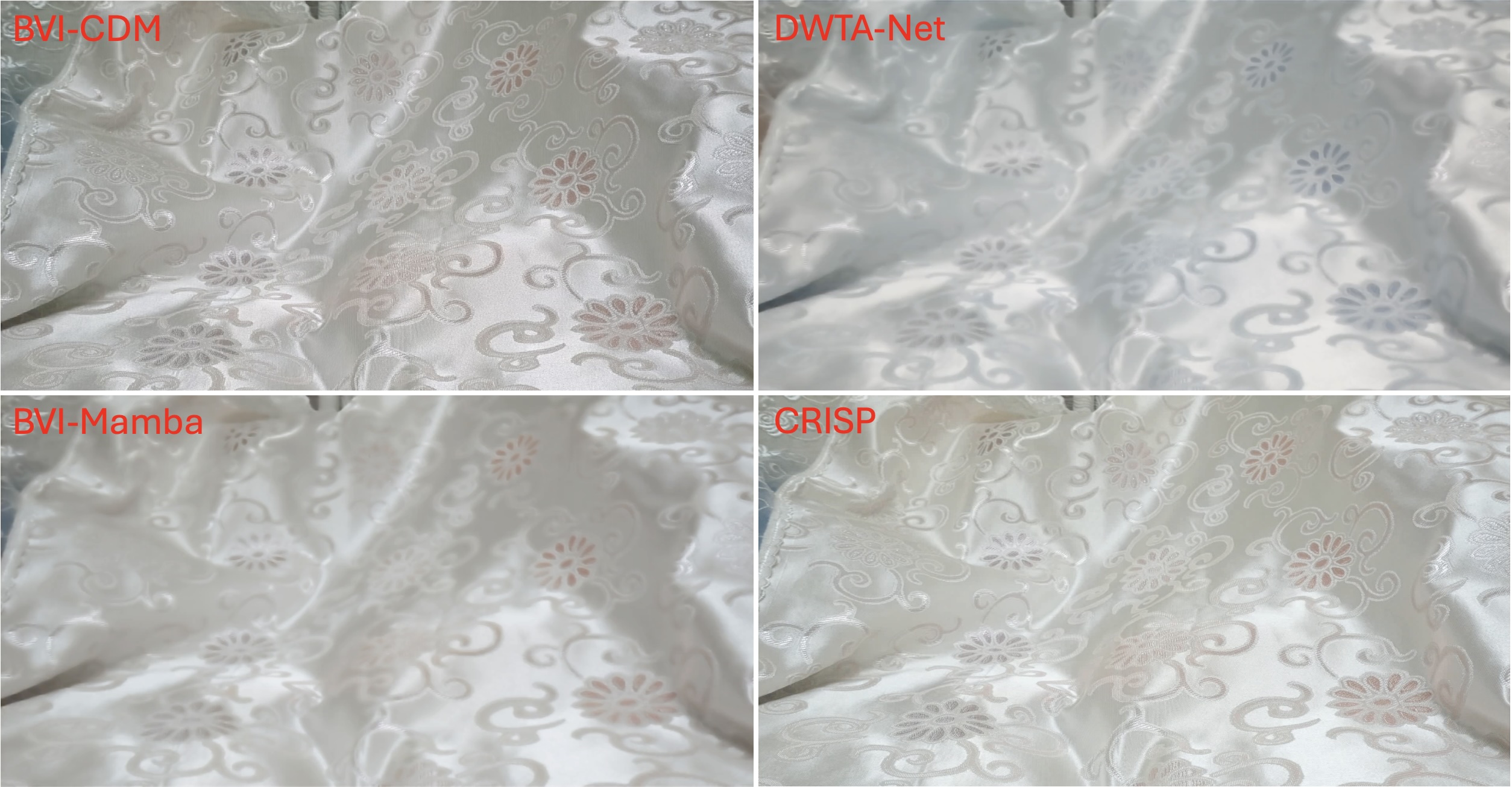}
    \caption{\textit{Additional qualitative comparison on the DID dataset. Clockwise from the top left: BVI-CDM, DWTA-Net, BVI-Mamba, and our CRISP method.}}
    \label{fig:DIDresults}
\end{figure}

\autoref{fig:DIDresults} shows additional results on the DID dataset. Although DWTA-Net achieves the best LPIPS score, its output is blurry. BVI-Mamba also produces blurry results, as this particular low-light scene contains substantial noise on the reflective surface. Both BVI-CDM and our CRISP method perform well in this scene. We note that the preferred illumination level is subjective and can be adjusted during post-production grading, whereas denoising and deblurring are much more difficult to perform manually and therefore benefit more from AI-based methods. % The results demonstrate that the proposed method achieves balanced restoration, improving visibility while preserving local structure and avoiding over-enhancement artifacts.

% --------------------------------------------------

% -----------------------------------------------------------
\subsection{Ablation Study}

To assess the contribution of each component, ablation experiments were conducted by removing individual modules or training objectives from the full system. \autoref{tab:ablation_study} reports the effect of removing DINOv3 guidance, channel-wise multi-head attention, CIELAB chroma supervision, and curriculum training.

\subsubsection{Semantic component analysis.}
\begin{figure*}[t]
    \centering
    \begin{minipage}[t]{0.24\textwidth}
        \centering
        \includegraphics[width=\linewidth]{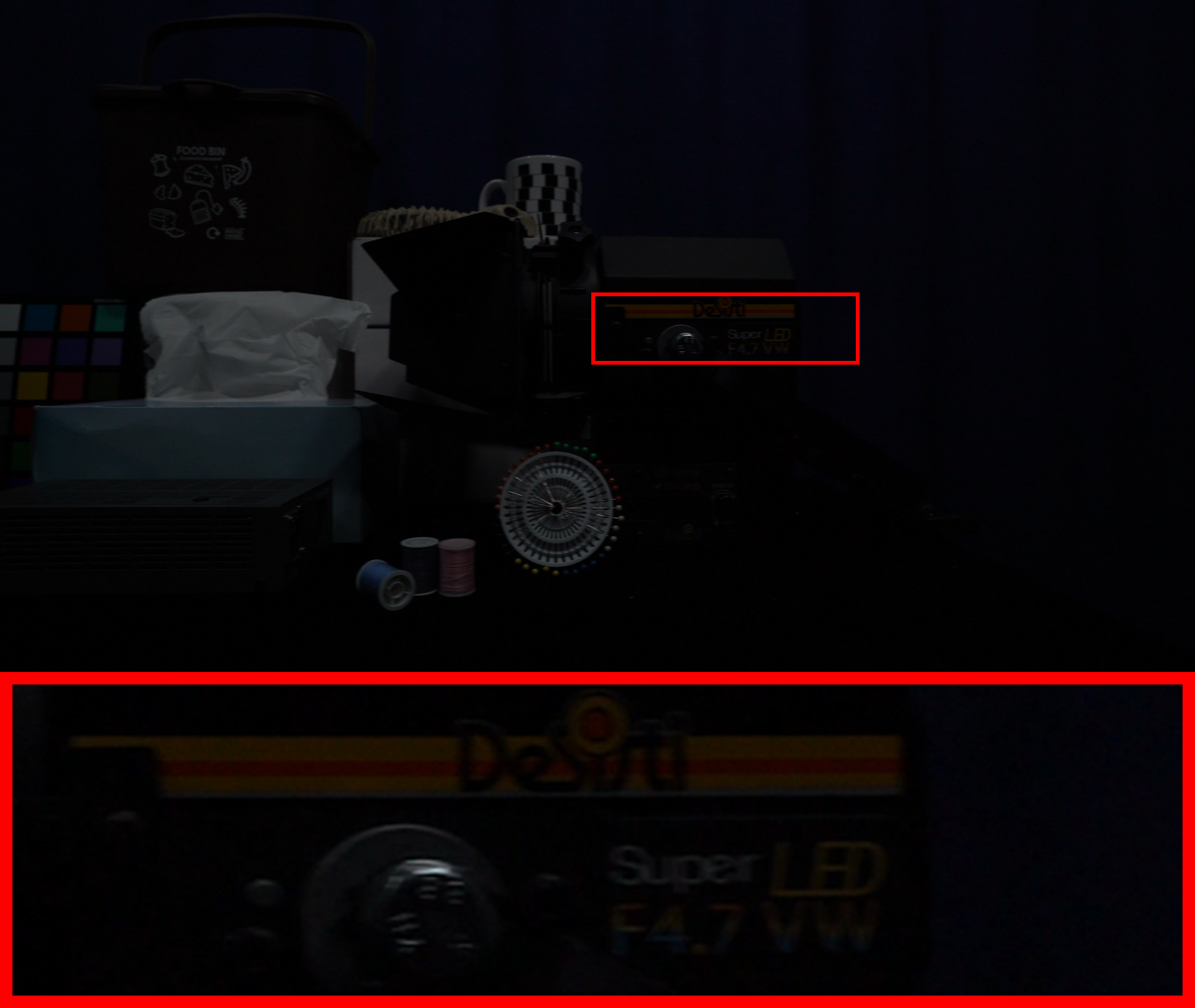}\\
        {\scriptsize Input}
    \end{minipage}%
    \begin{minipage}[t]{0.24\textwidth}
        \centering
        \includegraphics[width=\linewidth]{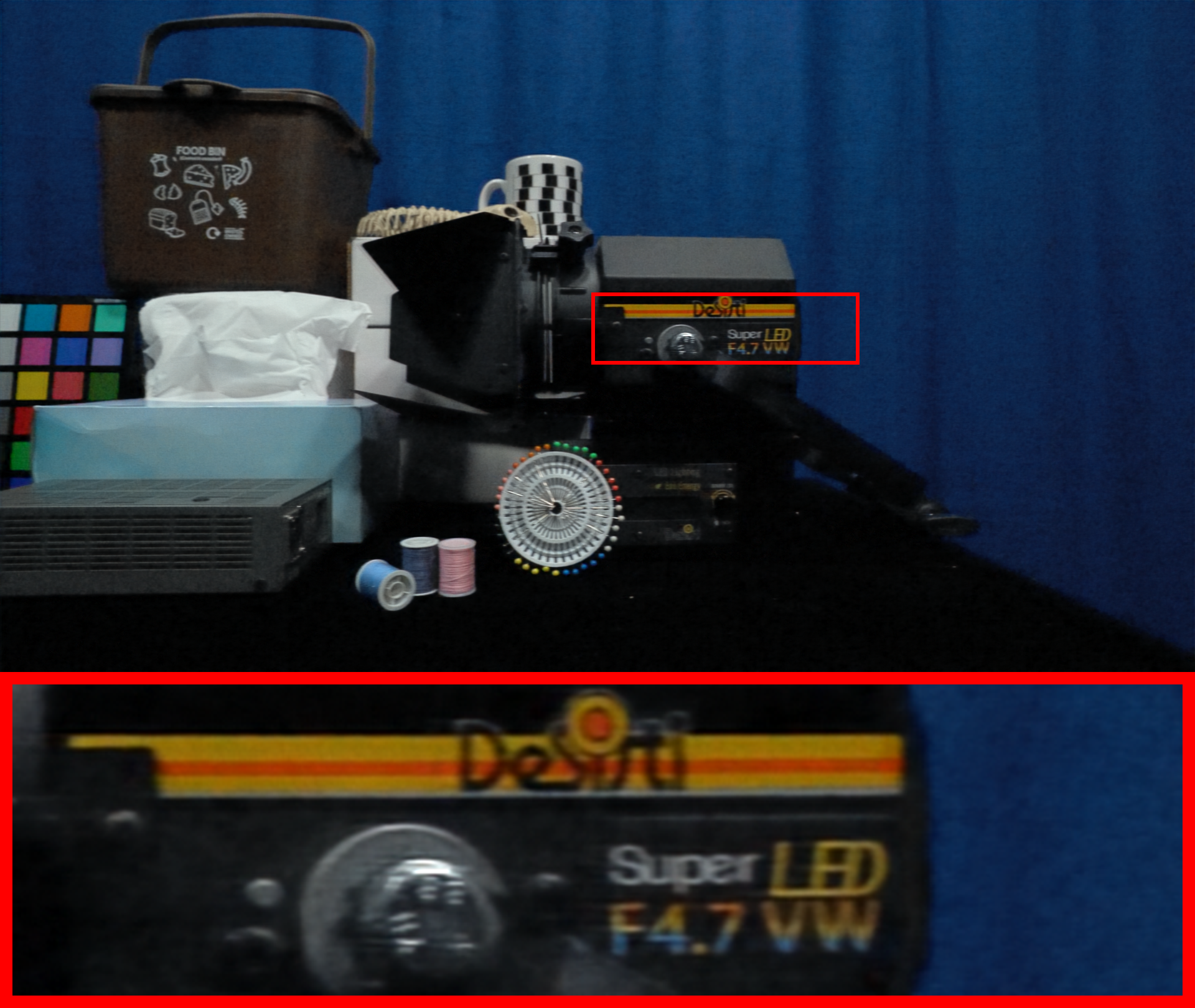}\\
        {\scriptsize w/o DINO}
    \end{minipage}%
    \begin{minipage}[t]{0.24\textwidth}
        \centering
        \includegraphics[width=\linewidth]{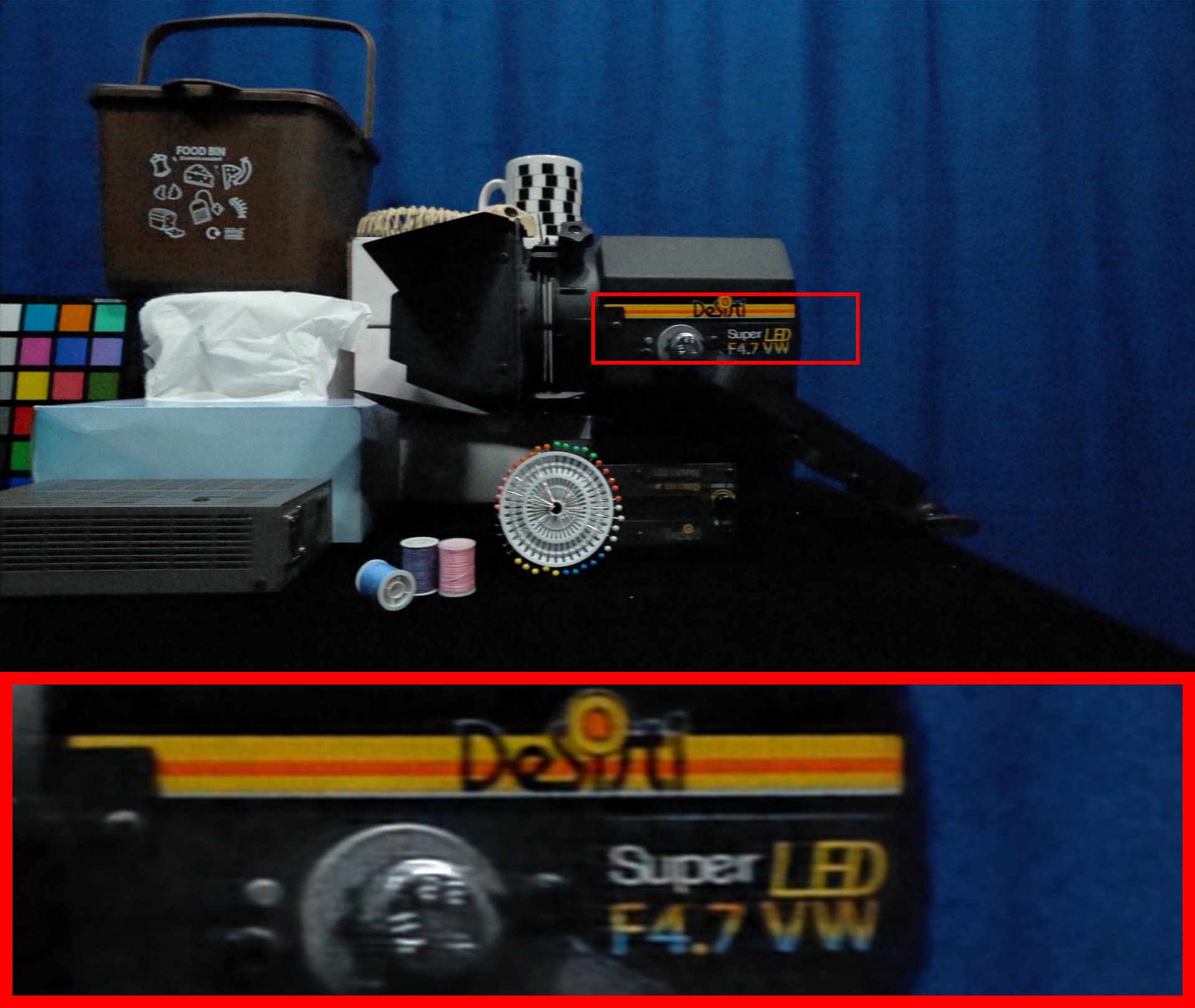}\\
        {\scriptsize w/ DINO}
    \end{minipage}%
    \begin{minipage}[t]{0.24\textwidth}
        \centering
        \includegraphics[width=\linewidth]{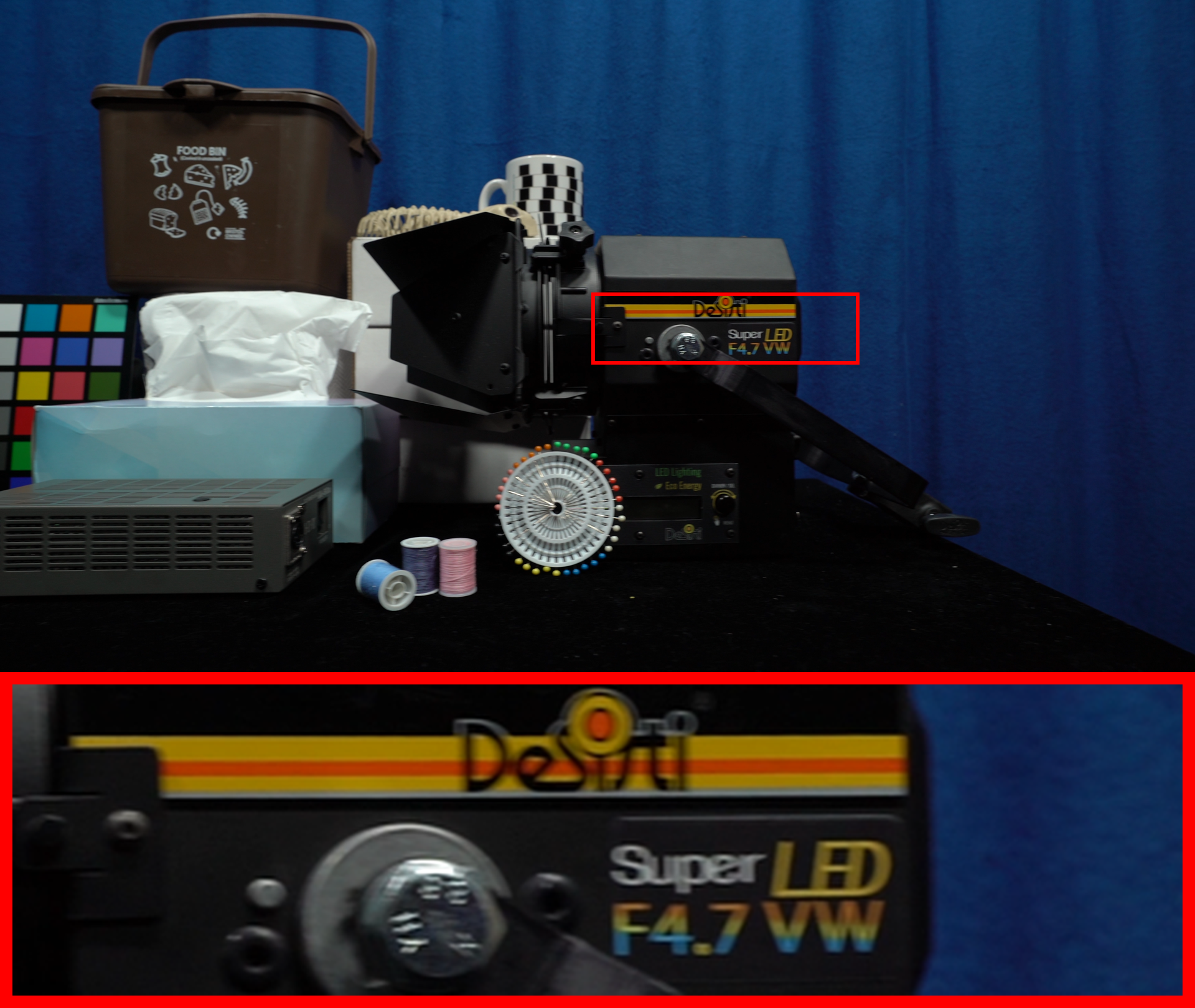}\\
        {\scriptsize GT}
    \end{minipage}
    %\vspace{4mm}
    \caption{\textit{Visual ablation of DINO guidance on the BVI-RLV dataset. The highlighted crop compares the model without DINO guidance against the DINO-guided model and ground truth.}}
    \label{fig:dino_ablation_visual}
\end{figure*}

Removing DINOv3 guidance slightly reduces PSNR (from 32.7768 to 32.7499), while causing a larger degradation in SSIM (from 0.9229 to 0.9154) and LPIPS (from 0.0984 to 0.1075). Semantic conditioning therefore improves structural and perceptual quality rather than pixel-wise accuracy. \autoref{fig:dino_ablation_visual} shows the corresponding loss of fine text and boundary sharpness when it is removed. As this frozen branch adds 31M non-trainable parameters (\autoref{sec:efficiency}), we treat it as an optional conditioning stage: the DINO-free configuration remains the more efficient operating point.

\subsubsection{Framewise restoration component analysis.}
We replace the channel-wise multi-head attention in the multi-scale encoder with conventional spatial attention. This results in a slight performance degradation, with PSNR decreasing from 32.78 to 32.73, SSIM from 0.923 to 0.913, and LPIPS increasing from 0.098 to 0.107. The more pronounced drop in SSIM and LPIPS suggests that the proposed formulation contributes to improved local structure preservation and perceptual consistency prior to recurrent temporal fusion.

\subsubsection{Loss function analysis.}
Removing chroma supervision decreases PSNR (from 32.7768 to 32.6021) and SSIM (from 0.9229 to 0.9188), while slightly improving LPIPS (from 0.0984 to 0.0961). This indicates that the chroma loss primarily acts as a color regularizer, improving reconstruction fidelity and structural consistency rather than perceptual similarity. Its main benefit is therefore more stable color recovery with reduced hue drift and desaturation.

\subsubsection{Training strategy analysis.}
Removing curriculum learning while retaining the same learning-rate schedule causes the largest degradation across all metrics: PSNR drops from 32.7768 to 31.4143, SSIM from 0.9229 to 0.8971, and LPIPS increases from 0.0984 to 0.1240. This demonstrates the benefit of the proposed staged training strategy, which improves optimization stability and enhances reconstruction quality, structural fidelity, and perceptual consistency.

\begin{figure*}[t]
    \centering
    \begin{minipage}[t]{0.19\textwidth}
        \centering
        \includegraphics[width=\linewidth]{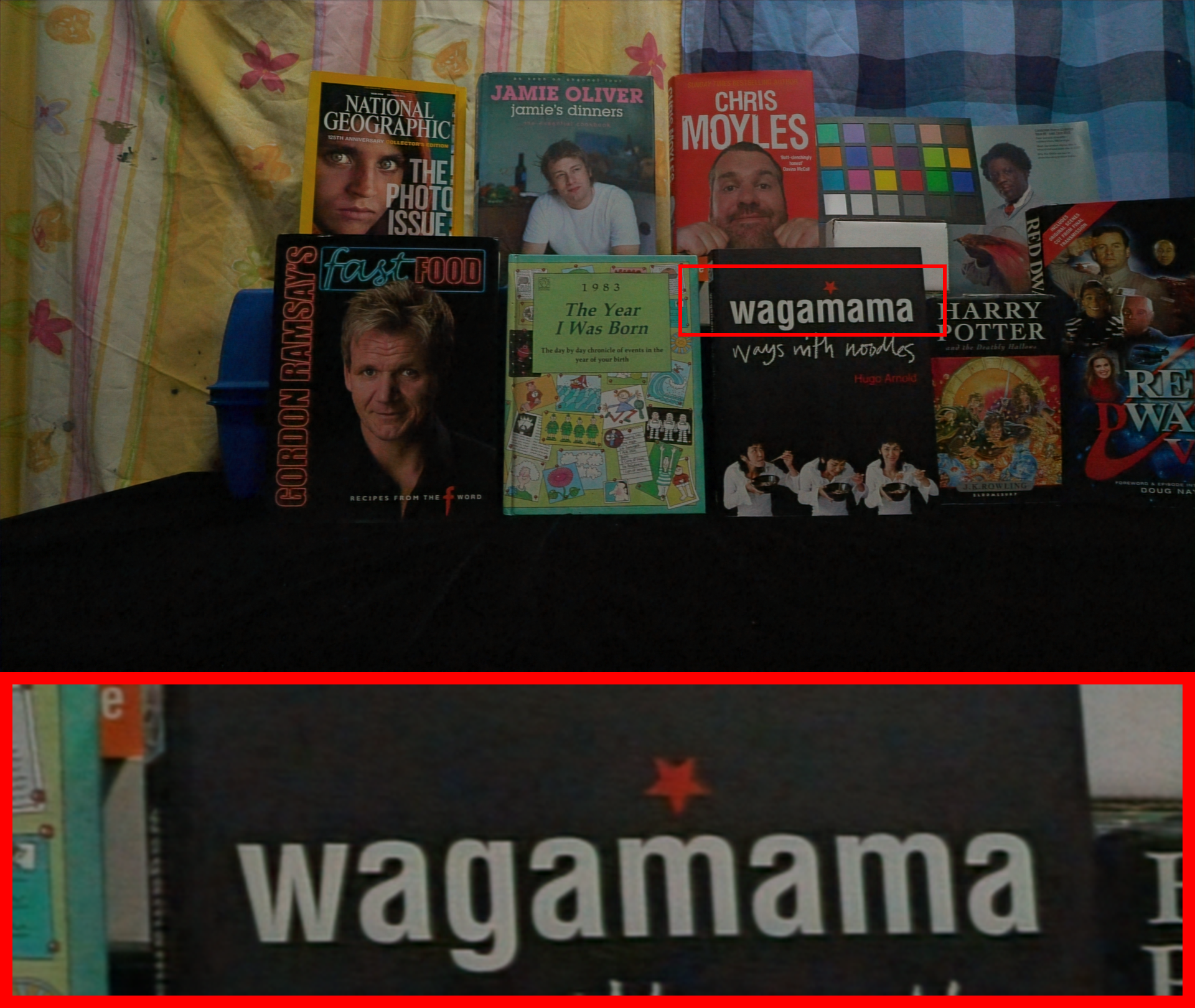}\\[-1mm]
        {\scriptsize Stage 1}
    \end{minipage}%
    \begin{minipage}[t]{0.19\textwidth}
        \centering
        \includegraphics[width=\linewidth]{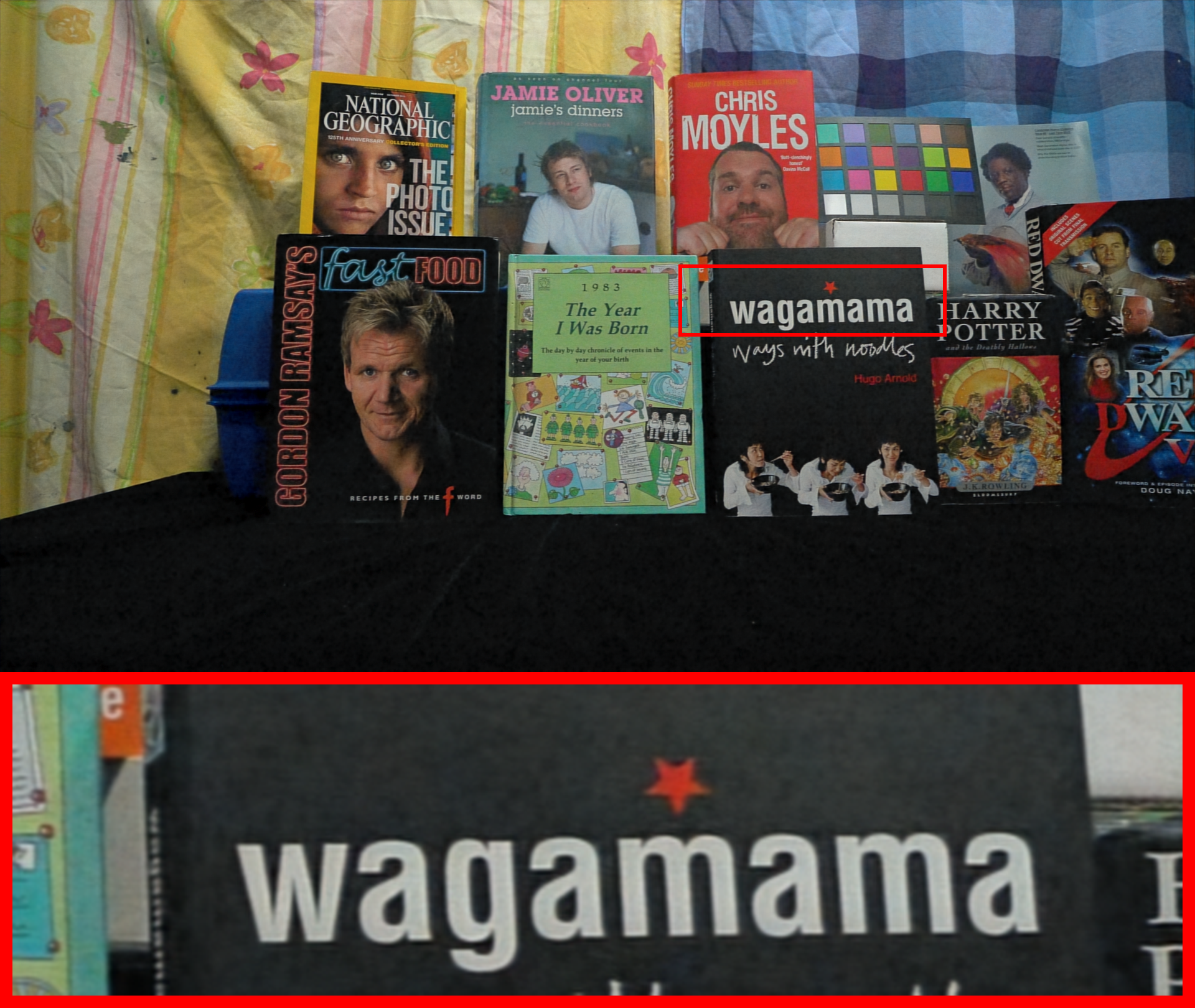}\\[-1mm]
        {\scriptsize Stage 2}
    \end{minipage}%
    \begin{minipage}[t]{0.19\textwidth}
        \centering
        \includegraphics[width=\linewidth]{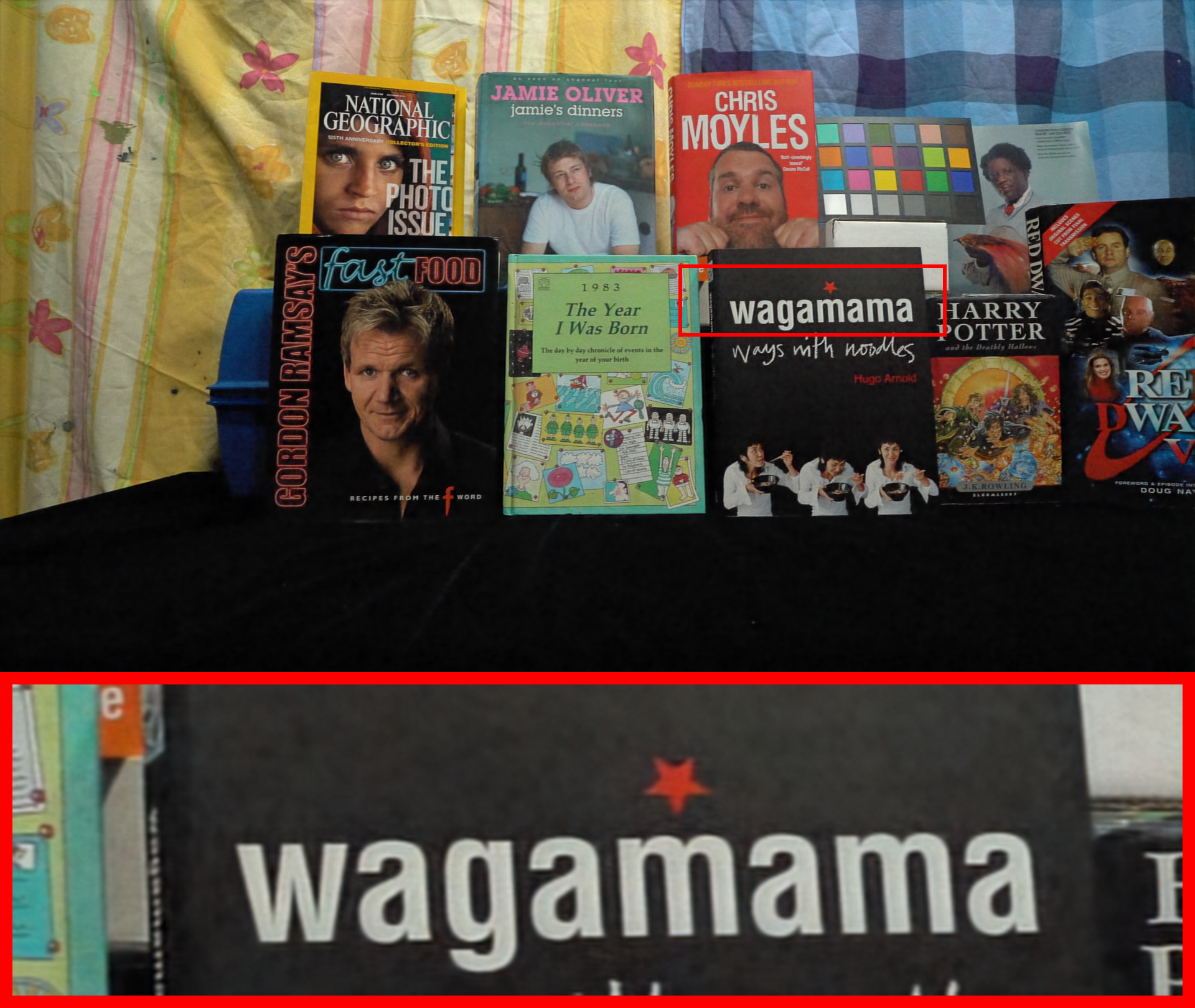}\\[-1mm]
        {\scriptsize Stage 3}
    \end{minipage}%
    \begin{minipage}[t]{0.19\textwidth}
        \centering
        \includegraphics[width=\linewidth]{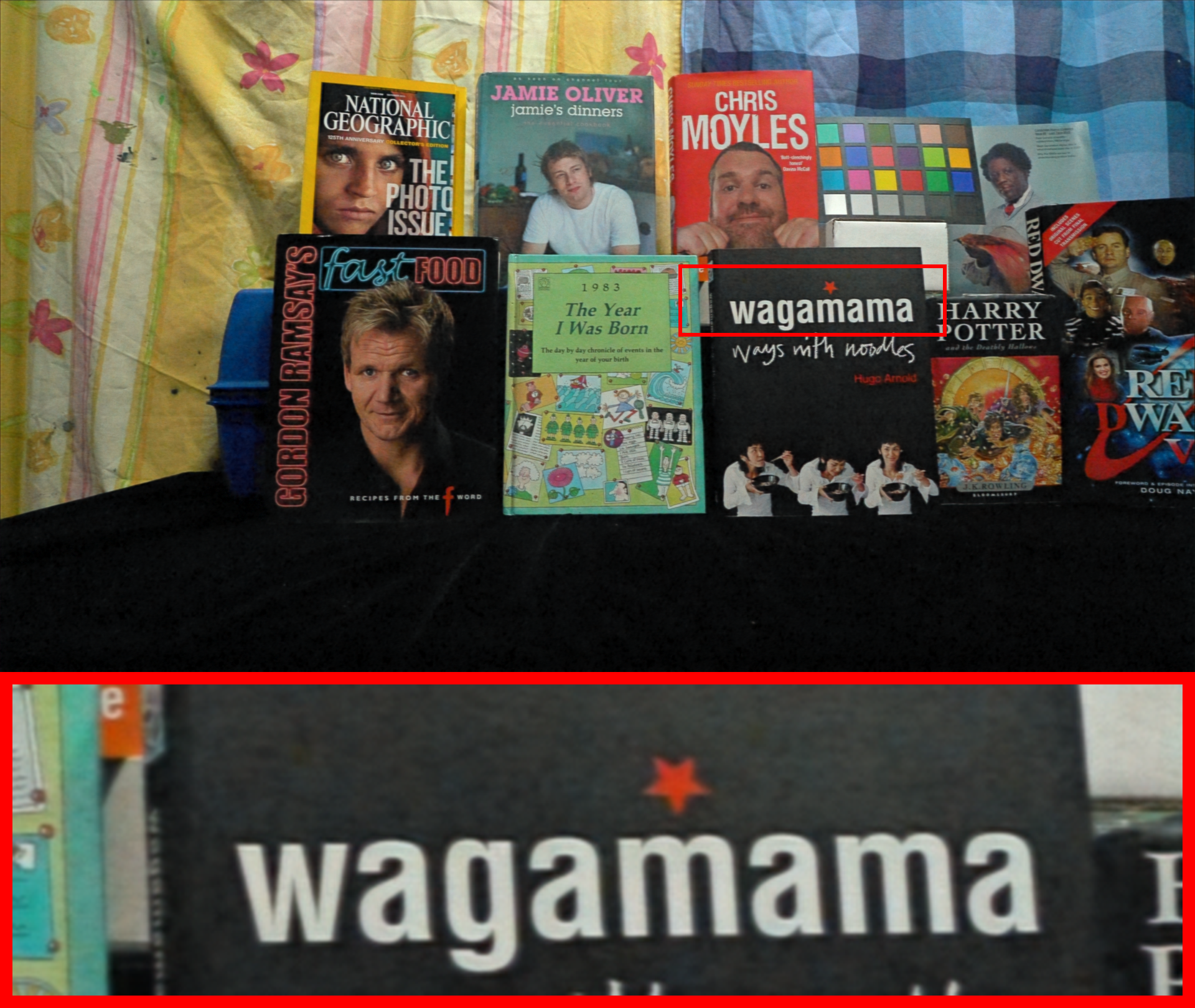}\\[-1mm]
        {\scriptsize w/o Curriculum}
    \end{minipage}%
    \begin{minipage}[t]{0.19\textwidth}
        \centering
        \includegraphics[width=\linewidth]{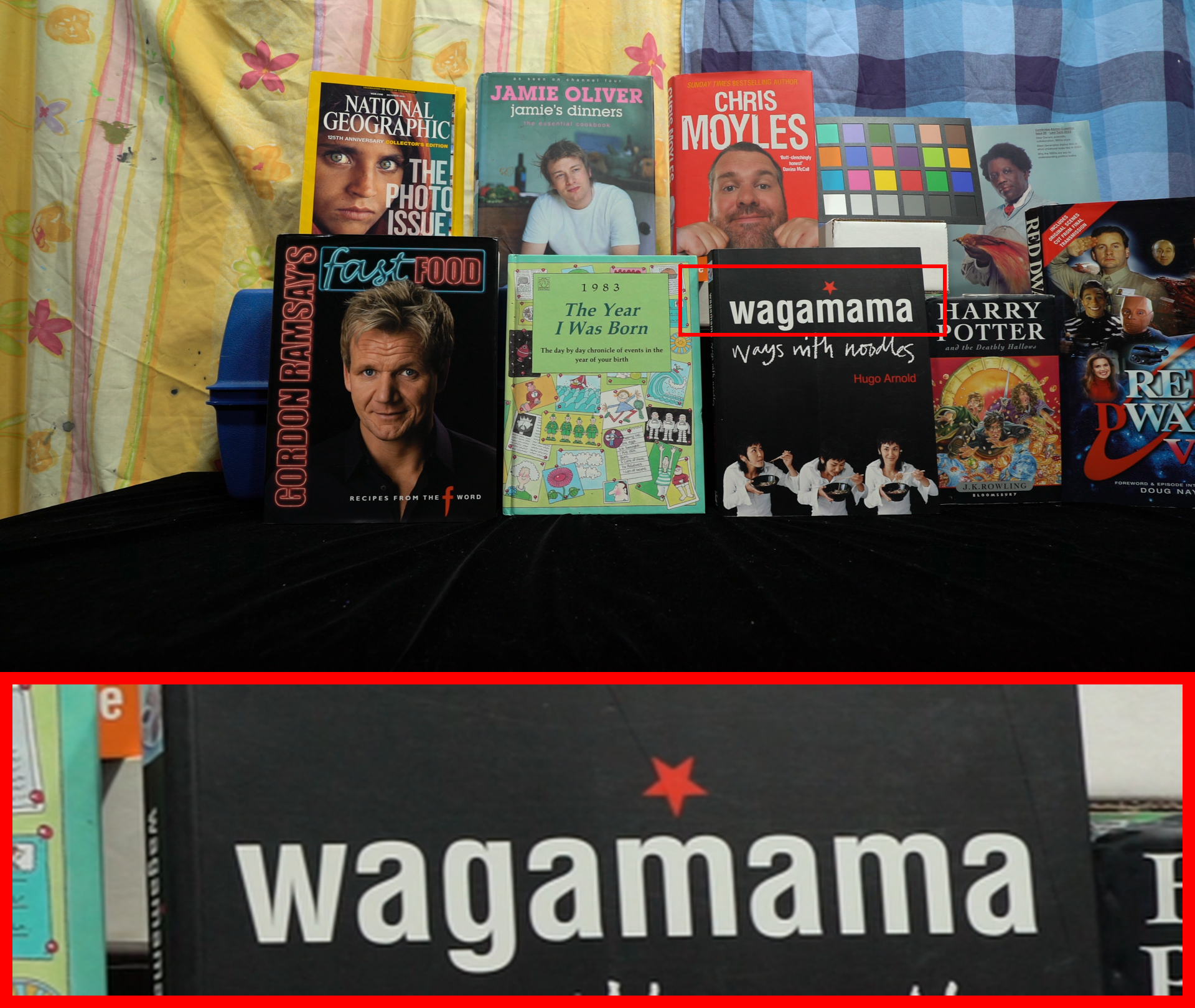}\\[-1mm]
        {\scriptsize GT}
    \end{minipage}

    %\vspace{4mm}
    \caption{\textit{Qualitative comparison between representative curriculum checkpoints and the model trained without curriculum learning on the BVI-RLV dataset. Stage 1, Stage 2 and Stage 3 are gradual restoration, brightness and color recovery, and full-reference refinement, respectively.}}
    \label{fig:curriculum_progression}
\end{figure*}

\autoref{fig:curriculum_progression}  further illustrates this progression. Stage~1 recovers the scene structure but remains dark with low local contrast. Stage~2 improves brightness and text legibility, while Stage~3 produces sharper text, more natural contrast, and color closer to the ground truth. Adding the temporal consistency constraint in Stage~4 does not degrade restoration quality, with PSNR, SSIM, and LPIPS changing only at the third decimal place when compared at the same training epoch. Without curriculum learning, brightness is similar but local contrast is less stable, highlighting the benefit of staged optimization for fine structural recovery.

    %\vspace{4mm}
   % \caption{Visual ablation of DINO guidance on the BVI-RLV dataset. The highlighted crop compares the model without DINO guidance against the DINO-guided model and ground truth.}

Notably, the ablation confirms that the illumination curriculum is the dominant factor, accounting for a 1.36\,dB PSNR gain, an order of magnitude larger than any architectural component, and obtained entirely at training time. Channel-wise attention and DINOv3 guidance mainly improve structural and perceptual quality, while chroma supervision supports color fidelity and reconstruction consistency, despite not producing the lowest LPIPS score.

\subsection{Complexity Analysis}
\label{sec:efficiency}
Complexity was measured using giga multiply-accumulate operations (GMACs) at 512$\times$512, while memory and FPS were evaluated at 1920$\times$1080. As shown in \autoref{tab:complexity}, compared with ReMATF (4.13M parameters, 164.82 GMACs), the DINO-free backbone requires only 2.92M parameters and 121.88 GMACs. Adding frozen DINOv3 increases complexity to 33.91M parameters and 194.49 GMACs, with 8.7GB memory and 0.90 FPS at Full HD resolution. This yields an accuracy--complexity trade-off. The lightweight backbone is more efficient than ReMATF, while DINOv3 trades throughput for the structural gains reported in \autoref{tab:ablation_study}.

\begin{table}
    \centering
    \caption{Ablation study on BVI-RLV dataset.}
   % \resizebox{\columnwidth}{!}{
    \begin{tabular}{l|ccc}
        \toprule
        {Configuration} & {PSNR $\uparrow$} & {SSIM $\uparrow$} & {LPIPS $\downarrow$} \\
        \midrule
        w/o DINO guidance & {32.7499} & 0.9154 & 0.1075 \\
        channel-wise $\rightarrow$ spatial-wise attentions & 32.7339 & 0.9134 & 0.1066 \\
        w/o Chroma loss & 32.6021 & 0.9188 & {0.0961} \\
        w/o Curriculum training & 31.4143 & 0.8971 & 0.1240 \\
        \hline
        Full model & 32.7768 & {0.9229} & 0.0984 \\
        \bottomrule
    \end{tabular}
   % }
    \label{tab:ablation_study} 
\end{table}

\begin{table}
    \centering
    \caption{Complexity analysis at $512\times512$ and $1920\times1080$.} 
   % \resizebox{\columnwidth}{!}{
    \begin{tabular}{lcccc}
        \toprule
        & \multicolumn{2}{c}{512$\times$512} & \multicolumn{2}{c}{1920$\times$1080} \\
        \cmidrule(lr){2-3} \cmidrule(lr){4-5}
        Architecture & Params (M)$\downarrow$ & GMACs$\downarrow$ & MEM$\downarrow$ & FPS$\uparrow$ \\
        \midrule
        BVI-Mamba~\cite{huang2025bvimamba} & 10.00 & 175.31 & - & - \\
        ReMATF~\cite{liu2026rematf} & 4.13 & 164.82 & - & - \\
        \midrule
        Ours w/o DINOv3 & 2.92 & 121.88 & 6.51\,GB & 1.13 \\
        Ours Full Model & 33.91 & 194.49 & 8.70\,GB & 0.90 \\
        \bottomrule
    \end{tabular}
  %  }
    
    \label{tab:complexity}
\end{table}

%-------------------------------------------------------------------------
\section{Conclusion}

We proposed \textbf{CRISP}, a semantic-guided recurrent framework for low-light video enhancement that jointly improves brightness, color fidelity, structural recovery, and temporal coherence. Combining two-frame recurrence, curriculum learning, frozen DINOv3 semantic priors, and chroma-aware supervision, CRISP enables stable optimization and effective temporal aggregation under severe degradation. Experiments on the DID, BVI-RLV and OLiVES benchmarks demonstrate strong quantitative and qualitative performance. The DINO-free configuration uses fewer parameters and operations than the selected baselines, while optional DINOv3 conditioning improves structural and perceptual metrics at additional computational cost. The full model achieves 0.90 FPS at $1920\times1080$ on the reported setup. Future work will explore advanced motion modeling for improved robustness in complex dynamic scenes.

%% Loading bibliography style file
%\bibliographystyle{model1-num-names}
\bibliographystyle{unsrt}

% Loading bibliography database
\bibliography{cas-refs}

% Biography
%\bio{}
% Here goes the biography details.
%\endbio

%\bio{pic1}
% Here goes the biography details.
%\endbio

\end{document}